\documentclass{article}

\PassOptionsToPackage{numbers, sort}{natbib}
\usepackage[main, final]{neurips_2026}
\usepackage{makecell}

\usepackage[utf8]{inputenc} 
\usepackage[T1]{fontenc}    
\usepackage{url}            
\usepackage{booktabs}       
\usepackage{amsfonts}       
\usepackage{multirow}       
\usepackage{amsmath}
\usepackage{algorithm}
\usepackage{algorithmic}
\usepackage{bm}
\usepackage{caption}
\usepackage{mathtools}
\usepackage[cal=cm]{mathalpha} 
\usepackage{wrapfig}
\usepackage{enumitem}
\usepackage{nicefrac}       
\usepackage{microtype}      
\usepackage{xcolor}         
\usepackage{graphicx}
\usepackage{multibib}
\usepackage[normalem]{ulem}
\usepackage[compact]{titlesec}
\usepackage[table]{xcolor}
\usepackage{array}
\usepackage[most]{tcolorbox}
\usepackage{tabularx}
\usepackage{threeparttable}
\usepackage{siunitx}
\usepackage{subcaption}
\usepackage{tabularray}
\usepackage[dvipsnames]{xcolor}

\definecolor{ourblue}{rgb}{0.21,0.49,0.74}
\usepackage[pagebackref,breaklinks,colorlinks,linkcolor=red,citecolor=ourblue]{hyperref}

\definecolor{softred}{HTML}{A85C5C}
\definecolor{softgreen}{HTML}{3F7F5F}
\definecolor{softgreenbg}{HTML}{EEF7F2}
\definecolor{headergray}{HTML}{F6F6F6}

\newtcbtheorem[number within=section]{research_question}{Research Question}{
  colback=gray!8,
  colframe=gray!60,
  coltitle=black,
  fonttitle=\bfseries,
  colbacktitle=gray!20,
  boxrule=0.6pt,
  arc=2pt,
  left=5pt,
  right=5pt,
  top=4pt,
  bottom=4pt,
}{ki}

\makeatletter
\def\RulerShowFour#1,#2,#3,#4,#5\@nil{%
  \ignorespaces#1 & \ignorespaces#2 & \ignorespaces#3%
}
\def\RulerSelectHideLengths#1;#2;#3;#4;#5;#6;#7\@nil{%
  \RulerShowFour#1,\@nil &
  \RulerShowFour#2,\@nil &
  \RulerShowFour#3,\@nil &
  \RulerShowFour#5,\@nil &
  \RulerShowFour#7,\@nil%
}
\newcommand{\RulerRowHideLengths}[3]{%
  \ifx\relax#3\relax
    #1 & \multicolumn{15}{c|}{} \\
  \else
    #1 &  \RulerSelectHideLengths#3\@nil \\
  \fi
}

\definecolor{lt}{RGB}{62, 141, 34}

\newtheorem{proposition}{Proposition}[section] 

\title{Hierarchical Sparse Attention Done Right: \\Toward Infinite Context Modeling}

\author{
\begin{tabular}{c}
Xiang Hu$^1$\thanks{Core Contributors. shawnxxxhu@tencent.com} \quad Xinyu Wei$^{2*}$ \quad Hao Gu$^{3*}$ \quad Minshen Zhang$^{4*}$ \quad Tian Liang$^1$ \\
Huayang Li$^1$ \quad Lei Zhu$^1$ \quad Yan Wang$^1$ \quad Sirui Han$^3$ \quad \\
Yushi Bai$^1$ \quad Kewei Tu$^2$ \quad Haitao Mi$^1$ \quad Leo Liang$^1$
\end{tabular}\\
$^1$Tencent HY Team,\quad $^2$ShanghaiTech University, \quad \\$^3$The Hong Kong University of Science and Technology \quad \\ $^4$University of California, San Diego \\
\url{https://github.com/Tencent-Hunyuan/HiLS-Attention}
}

\begin{document}

\maketitle


\begin{figure}[h!]



    
    \begin{subfigure}[b]{0.9\linewidth}
        \centering
        \includegraphics[width=\linewidth]{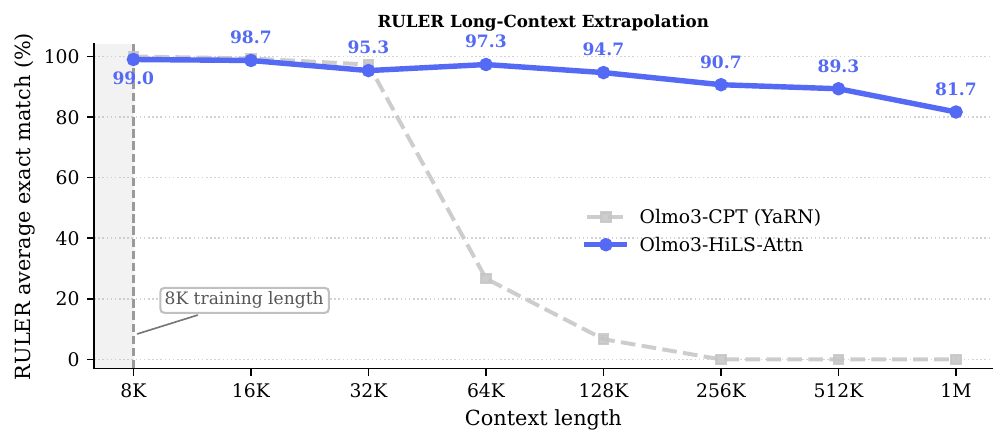}
        \caption{}
        \label{fig:ruler_extra}
    \end{subfigure}
    \begin{subfigure}[b]{0.4\linewidth}
        \centering
        \includegraphics[width=\textwidth]{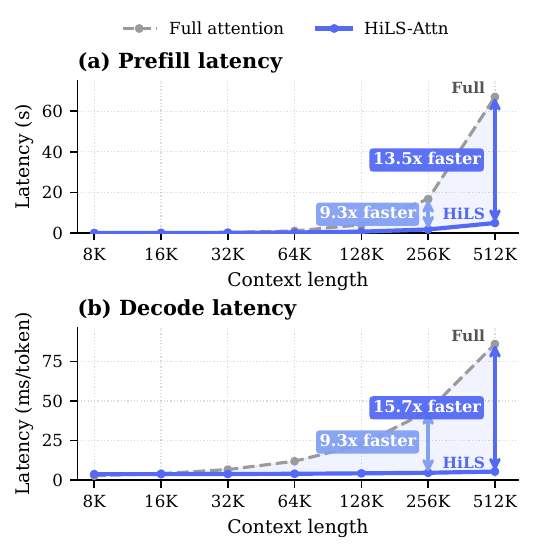}
        \caption{}
        \label{fig:efficiency_speedup}
    \end{subfigure}
    \begin{minipage}[b]{0.55\linewidth}
        \centering
        \begin{subfigure}{\linewidth}
            \centering
            \includegraphics[width=\linewidth]{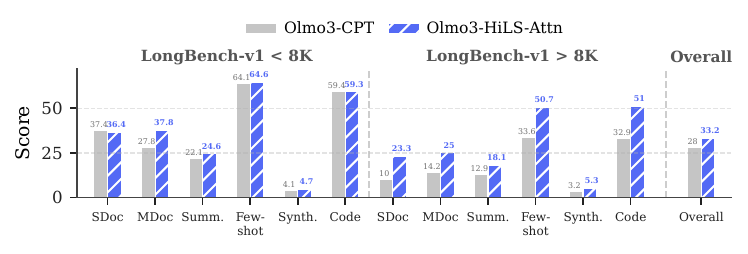}
            \caption{}
            \label{fig:longbench_v1}
        \end{subfigure}
        \begin{subfigure}{\linewidth}
            \centering
            \includegraphics[width=\linewidth]{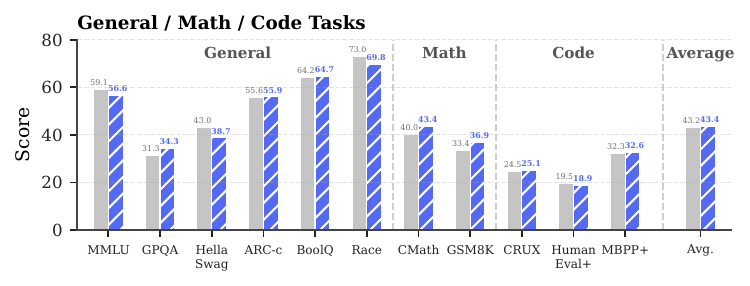}
            \caption{}
            \label{fig:general_task}
        \end{subfigure}
    \end{minipage}

    \caption{After only 50B continued-training tokens, HiLS-Attention inherits the capability of full attention while bringing two key advantages: \textit{strong ultra-long context extrapolation} beyond the YaRN-extended 4$\times$ length (Fig.~\ref{fig:ruler_extra}) and \textit{faster inference} (Fig.~\ref{fig:efficiency_speedup}) . Meanwhile, it preserves comparable performance for  short- and medium-context tasks, within both the original training length and the YaRN-extrapolated range (Fig.~\ref{fig:longbench_v1} \& \ref{fig:general_task}). 
    }
    \label{fig:main_results}
\end{figure}

\newpage
\begin{abstract}
Scaling modern large language models (LLMs) to long contexts is limited by the quadratic computation cost, and poor length extrapolation of dense attention. Chunk-wise sparse attention offers a promising alternative, but all existing methods fall short of full attention because of their inaccurate chunk selection.
We propose \textbf{Hi}erarchical \textbf{L}andmark \textbf{S}parse (HiLS) Attention, a chunk-wise sparse attention mechanism that learns chunk selection end-to-end under the language-modeling (LM) loss. HiLS factorizes attention hierarchically: each query performs attention independently with each retrieved chunk to extract chunk-specific information, and the resulting outputs are fused according to chunk retrieval scores. By incorporating retrieval scores into the forward attention computation, HiLS optimizes them directly with the LM loss, enabling end-to-end retrieval learning and native sparse training.
Experimental results in Fig~\ref{fig:main_results} show that HiLS-Attention achieves performance comparable to, and in some cases better than, full attention at in-domain context lengths. Meanwhile, HiLS-Attention extrapolates more than $64\times$ the training context length with 90\% retrieval accuracy, far beyond full attention. Moreover, existing full-attention models can be converted to HiLS-Attention with lightweight continued pretraining, preserving in-domain performance while acquiring ultra-long-context extrapolation. Together with its sparse KV access and computation, HiLS-Attention breaks the usual efficiency-performance trade-off, enabling long-context LLMs that are both more efficient and more effective on general long-context tasks than their full-attention counterparts.
\end{abstract}

\section{Introduction}

The ability to model and utilize long contexts has become a fundamental desired capability of modern Large Language Models (LLMs)~\cite{brown2020language,achiam2023gpt}. Consequently, scaling the context window has emerged as a key research frontier, underpinning a wide range of long-context applications such as long-horizon agentic tasks, complex reasoning, and large-scale information integration.
Despite recent progress~\cite{liu2024ringattention}, scaling context windows remains challenging for full attention due to its quadratic complexity, poor length extrapolation performance, and  KV cache cost that grows with the context length.

\begin{wrapfigure}{r}{0.42\textwidth}
\vspace{-6mm}
 \centering
\includegraphics[trim=0 0 0 0, clip, width=0.42\textwidth]{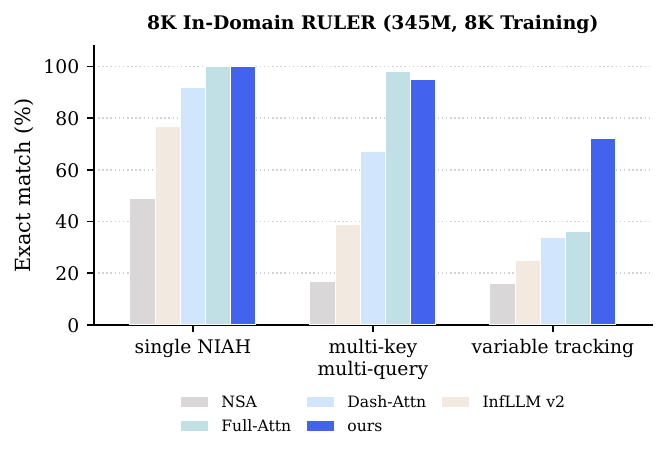}
\captionsetup{font=small}
\caption{In-context retrieval results.}
\label{fig:ruler_results}
\vspace{-6mm}
\end{wrapfigure}
Recently, chunk-wise sparse attention methods~\cite{hu2025efficient,yuan-etal-2025-native,DBLP:journals/corr/abs-2502-13189,huang2026nosanativeoffloadablesparse} provide a promising alternative. They selectively attend to relevant context chunks to maintain a constant computational cost, while dynamically swapping the corresponding KV caches into fast memory on demand to prevent memory explosion.
Despite their recent progress, no existing chunk-wise native sparse attention methods have achieved performance on par with full attention methods. While the performance gap may appear small for large-scale models on short-context tasks, it becomes pronounced in longer-context settings that demand precise in-context retrieval. This limitation is clearly exposed in parameter-constrained models, as evidenced by the substantial long-context retrieval gap in Fig.~\ref{fig:ruler_results}.
We argue that inaccurate chunk selection is the core challenge, which stems from weak chunk summaries and the lack of end-to-end optimization of the selection process. The nonparametric chunk summaries such as mean-pooling, have limited expressiveness and may lose crucial information.  
Although parametric summaries could be more expressive, existing methods use them only to score chunks for selection: after the hard top-$K$ chunk IDs are chosen, the summaries and chunk scores are discarded.
Consequently, the language-modeling (LM) loss cannot directly optimize the summaries or selection scores to suppress irrelevant chunks and promote chunks more useful for next-token prediction, leading to inaccurate chunk selection.

\begin{figure}[tb!]
    \centering
    \includegraphics[width=0.98\linewidth]{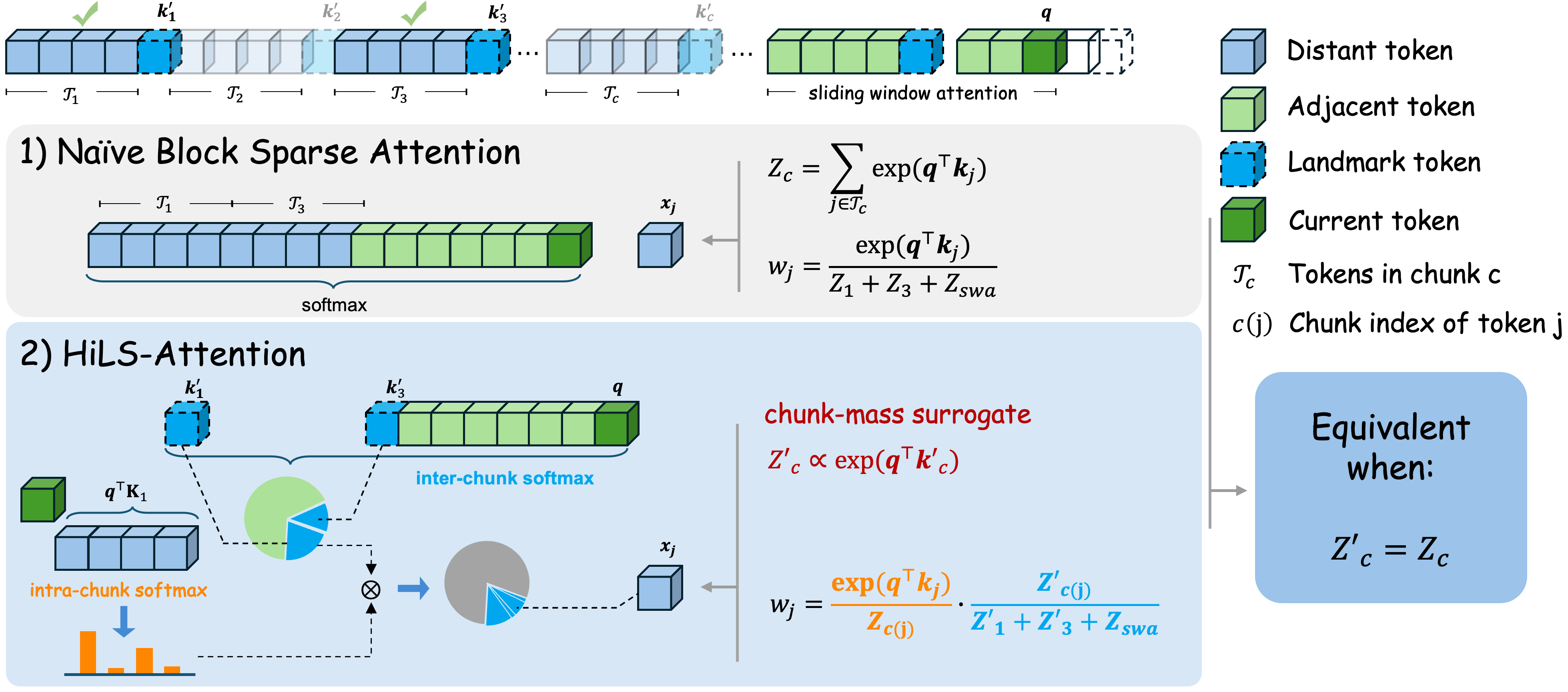}
    \caption{An overview of HiLS-Attention.
    We omit the scaling factor $\frac{1}{\sqrt{d}}$ for simplicity. Naive block sparse attention selects the top-$K$ chunks by their exact mass $Z_c$, e.g., chunks 1 and 3 when $K=2$, but computing all $Z_c$ requires a full QK computation. HiLS-Attention instead uses compressed chunk keys $k'_c$ to efficiently estimate a chunk-mass surrogate $Z'_c\propto \exp(q^\top k'_c)$. It factorizes attention into two stages: an \textbf{inter-chunk softmax}, which specifies the total attention mass assigned to each chunk, and an \textbf{intra-chunk softmax}, which distributes each chunk’s attention mass among its tokens. Since $Z'_c$ parameterizes the forward attention weights, gradients from the next-token prediction loss can be directly backpropagated to the compressed key $k'_c$, enabling end-to-end learning.}
    \label{fig:HilS-Attn}
\end{figure}

This observation motivates two desiderata: chunk summaries should be trainable end-to-end with the LM loss, and expressive enough to capture the chunk-level importance induced by full attention.
To explore this, we start from the naive Block Sparse Attention (BSA), which computes full attention, aggregates token-level attention mass within each chunk, and selects the top-$K$ chunks with the largest mass. 
Although naive BSA requires full attention computation and offers no computational savings, it yields a full-attention-derived chunk selection pattern. We use this pattern as a starting point to derive \textbf{HiLS}-Attention, an end-to-end learnable sparse attention mechanism with sufficient expressiveness to capture such chunk-level mass.
First, to estimate chunk mass without computing full attention, we append a special \textit{landmark token}~\cite{mohtashami2023random} to each chunk, from which we derive an expressive \textit{chunk summary key}. The query-summary-key scoring operation follows the first-order Taylor expansion of the full-attention induced chunk mass, thereby forming a learnable chunk-mass surrogate.
Second, HiLS makes this retrieval process end-to-end learnable by using the surrogate scores as part of the forward attention weights, via the hierarchical factorization illustrated in Fig.~\ref{fig:HilS-Attn}.
In this factorization, attention mass is first allocated across retrieved chunks and then distributed among tokens within each chunk. This avoids the quadratic full-attention pass required by naive BSA, thereby substantially reducing the computational cost during both training and inference. As a result, block selection becomes end-to-end learnable under the LM objective, allowing HiLS to assign higher scores to chunks that are more useful for prediction and suppress irrelevant ones.

To validate HiLS-Attention's alignment with naive BSA and full attention, and more importantly, assess its performance for long-context modeling, we conduct comprehensive experiments across model scales from 345M to 7B parameters, covering perplexity evaluation, short-context benchmarks, and long-context benchmarks. 
At the 345M scale, HiLS-Attention achieves comparable perplexity to full attention and better in-domain RULER~\cite{hsieh2024ruler} performance. Remarkably, despite   pretrained with only  8K context length, it extrapolates to 4M context length while maintaining over 90\% accuracy on needle-in-a-haystack retrieval, corresponding to a $512\times$ length extrapolation.
We further find that the advantage of HiLS-Attention becomes more pronounced with  256K training context length. On a challenging variable-tracking task, it improves over full attention by up to 50\%. At the 7B scale, we can convert a full-attention model to HiLS-Attention with only 50B tokens of continual training, preserving short-context performance while substantially outperforming full-attention baselines on LongBench~\cite{bai2024longbench} and exceeding the YaRN-extended~\cite{peng2024yarn} base model.
These results suggest that \textit{HiLS-Attention can not only match but also surpass these baselines in multiple settings.} 

To our best knowledge, our work is the first to provide strong empirical evidence that native sparse attention can achieve both superior long-context performance and more efficient long-context inference. Together, its strong in-domain performance, superior length extrapolation, and efficient long-context inference position HiLS-Attention as a promising alternative to full attention and a core building block for future ultra-long context modeling. 

In summary, our key contributions are as follows:
\begin{itemize}[leftmargin=*,itemsep=0.2em,topsep=0.3em]
\item We propose {HiLS-Attention}, a native sparse attention mechanism based on a hierarchical softmax, enabling end-to-end trainable sparse retrieval.
\item We connect HiLS-Attention to naive BSA from an expressiveness perspective, showing that an effective chunk summary should be mathematically aligned with the first-order Taylor expansion of the full-attention-induced chunk mass, thereby providing sufficient representational capacity for accurate block selection.
\item Our extensive experiments show that full-attention models can be cost-effectively converted to HiLS-Attention, preserving short-context performance while surpassing full attention in both in-domain long-context tasks and ultra-long-context extrapolation.
\end{itemize}

\section{Preliminary}

In this section, we review the formulation of naive Block Sparse Attention (BSA) and analyze why existing methods fail to preserve the fidelity of chunk selection.

\subsection{Naive Block Sparse Attention}\label{sec:naive_bsa}
Given an input token sequence $\boldsymbol{x} = \{x_1, x_2, \ldots, x_N\}$, we partition it into non-overlapping chunks of a uniform size $S$, where the $j$-th token belongs to the $c(j)$-th chunk ($c(j) = \lfloor j / S \rfloor$). 
With BSA, the query token attends to two distinct sequence segments: a local sliding window containing the corresponding token and $K$ globally selected distant chunks outside this window.

To avoid overlap between the local window and the retrieved distant chunks, we align the local window with the chunk partition by rounding its left boundary down to the nearest chunk boundary: given a window size $W$, the aligned left boundary at position $i$ is $\ell(i)=\left\lfloor \frac{i-W+1}{S} \right\rfloor S$.
Consequently, at time step i, the candidate chunks indices available for retrieval from the historical context form an indices set $\mathcal{C}_i = \{0, 1, \ldots, \frac{\ell(i)}{S}-1\}$. 

Let $\mathbf{q}_i, \mathbf{k}_j, \mathbf{v}_j \in \mathbb{R}^d$ denote the standard query of $x_i$, and key and value vectors of $x_j$ with dimension $d$. The token-to-token attention logit is defined as $s_{i,j} = \frac{\mathbf{q}_i^\top \mathbf{k}_j}{\sqrt{d}}$. In the full-attention setting, 
we define the \textit{chunk mass} of a token group as the sum of token-level exponentiated attention logits. 
Thus, the chunk masses assigned to the local sliding-window attention (SWA)  region and to a distant chunk $c \in \mathcal{C}_i$ for token position $i$ are respectively given by
\begin{equation}
\small
\label{eqn:z_bsa}
 Z_{i, \text{swa}} = \sum_{j=\ell(i)}^{i} \exp(s_{i,j}), \quad 
     Z_{i,c} = \sum_{j\in\mathcal{T}_c} \exp(s_{i,j}),
\end{equation}
Since selecting chunks according to normalized attention mass is equivalent to selecting them according to $Z_{i,c}$, naive BSA chooses
\begin{equation}
\small
\label{eq:naive_bsa_select}
    \bm{\mathcal{I}}_i
    =
    \left\{
    c \in \mathcal{C}_i
    \mid
    \operatorname{rank}_{\downarrow}(
    Z_{i,c} 
    ) < K
    \right\},
\end{equation}
where $\operatorname{rank}_{\downarrow}(Z_{i,c})$ ranks chunks in $\mathcal{C}_i$ in descending order of $Z_{i,c}$, so $\bm{\mathcal{I}}_i$ contains the top-$K$ chunk indices.
The query then attends only to selected tokens, i.e., tokens from the selected chunks and the local sliding window. Thus, we have
\begin{equation}
\small
\label{eq:naive_bsa_weight}
\begin{aligned}
    &\mathcal{Z}_i =
    \sum_{c \in \mathcal{I}_i} Z_{i,c}
    +
    Z_{i,\mathrm{swa}}, \\
    &w_{i,j} =
    \begin{cases}
    \displaystyle
    \frac{\exp(s_{i,j})}{\mathcal{Z}_i},
    & j \text{ is selected} \\
    0,
    & \text{otherwise}
    \end{cases},\quad \mathbf{o}_{i} = \sum_{j \le i} w_{i,j} \mathbf{v}_j,
\end{aligned}
\end{equation}
where the softmax weights are normalized only over the selected tokens, while all unselected tokens receive zero weight.

\subsection{Expressiveness Limitations of Existing Methods}
\label{sec:exp_limit}

While naive BSA yields exact chunk selection, computing the precise $Z_{i,c}$ requires evaluating all token-level logits $s_{i,j}$ within chunk $c$, which diminishes the computational advantages of attention sparsity during training.
Therefore, an ideal sparse attention mechanism should estimate the chunk mass without explicitly computing all token-to-token scores. In particular, if one could construct a chunk-level summary key $\mathbf{k}'_c$ such that
\begin{equation}
\exp\left(\frac{\mathbf{q}_i^\top \mathbf{k}'_c}{\sqrt{d}}\right)
\approx
Z_{i,c}
=
\sum_{j\in \mathcal{T}_c}\exp(s_{i,j})
\Longleftrightarrow
\frac{\mathbf{q}_i^\top \mathbf{k}'_c}{\sqrt{d}}
\approx
\log Z_{i,c},
\end{equation}
then chunk selection could be performed efficiently using only chunk-level representations. However, the target $\log Z_{i,c}$ is a LogSumExp function, whose behavior depends on the logit distribution within the chunk:
\begin{equation}
\small
\label{eq:two_regime}
\log Z_{i,c}
=
\log \sum_{j\in\mathcal{T}_c}\exp(s_{i,j})
\approx
\begin{cases}
\operatorname{mean}_{j\in\mathcal{T}_c}(s_{i,j})+\log S,
& \text{if logits are nearly uniform}, \\[2mm]
\max_{j\in\mathcal{T}_c}(s_{i,j}),
& \text{if one logit dominates the others}.
\end{cases}
\end{equation}
A detailed derivation is provided in Appendix~\ref{apdx:logsumexp_two_regime}.

Most sparse attention methods approximate such chunk-level representations using mean-pooled keys~\cite{yuan-etal-2025-native,DBLP:journals/corr/abs-2502-13189}. Specifically, by setting $\mathbf{k}'_c$ as the average key within chunk $c$, we have
\begin{equation}
\small
\begin{aligned}
&\mathbf{k}'_c=\frac{1}{S}\sum_{j\in\mathcal{T}_c}\mathbf{k}_j,\quad \frac{\mathbf{q}_i^\top\mathbf{k}'_c}{\sqrt{d}}
=
\frac{\mathbf{q}_i^\top\left(\frac{1}{S}\sum_{j\in\mathcal{T}_c}\mathbf{k}_j\right)}{\sqrt{d}}
=
\frac{1}{S}\sum_{j\in\mathcal{T}_c}
\frac{\mathbf{q}_i^\top\mathbf{k}_j}{\sqrt{d}}
=
\operatorname{mean}_{j\in\mathcal{T}_c}(s_{i,j}).
\end{aligned}
\end{equation}
Thus, the resulting chunk score corresponds to the mean of token-level logits.

This reveals a fundamental limitation: mean logits are effective only when logits are nearly uniform, whereas max logits are accurate only when a single token dominates. Since logit distributions vary across queries, heads, and data, neither mean nor max logits~\cite{lai2026minimaxsparseattention} can consistently represent chunk mass faithfully. This mismatch may alter chunk rankings and cause important chunks to be missed.
\section{Methodology}\label{sec:method}

The above analysis motivates us to replace non-parametric chunk summaries with learnable summaries for chunk selection. 
We propose HiLS-Attention, a fully differentiable sparse attention mechanism that replaces the chunk mass $Z_{i,c}$ with a surrogate $\hat{Z}_{i,c}$ computed from learned chunk summaries.
This raises two key research questions:
\begin{research_question}{}{}
\begin{itemize}[leftmargin=*,itemsep=0.2em,topsep=0.3em]
    \item \textbf{RQ1:} Can the LogSumExp chunk mass $\log Z_{i,c}$ be approximated in a tractable form using only the interaction between the query $\mathbf{q}_i$ and a compact chunk summary?
    \item \textbf{RQ2:} How can we learn $\hat{Z}_{i,c}$ end-to-end so that chunk selection is optimized jointly with the language-modeling loss?
\end{itemize}
\end{research_question}
\paragraph{Answering RQ1: A Linear Surrogate for Chunk Mass.}
To identify a tractable surrogate for the LogSumExp chunk mass, we first examine its Taylor expansion. This reveals a useful affine structure for chunk-level scoring, as formalized in the following proposition.
The detailed derivation is provided in Appendix~\ref{appdx:proof_of_logsumexp}.
\begin{proposition} \label{prop:logsumexp} (LogSumExp Linearization). 
For any chunk $c$, the LogSumExp of query-key logits can be linearized as:
\begin{equation}
\small
    \log \sum_{j \in \mathcal{T}_c} 
    \exp\left(\frac{\mathbf{q}^\top\mathbf{k}_j}{\sqrt d}\right) 
    \approx 
    \underbrace{\frac{\mathbf{q}^\top \mathbf{k}'_c}{\sqrt{d}} + b'_c}_{\mathrm{surrogate~score}},
    \label{eq:attn_lhsa}
\end{equation}
where $\mathbf{k}'_c$ and $b'_c$ are computed from a learnable query $\mathbf{q}'_c$ and $\mathbf{k}_i$ in chunk $c$, where $i \in [cS,(c+1)S)$; the concatenation of $\mathbf{k}_i$ is denoted by $\mathbf{K}_c \in \mathbb{R}^{S\times d}$. Specifically,
\begin{equation}
\small
    s_j=\frac{(\mathbf{q}'_{c})^{\top} \mathbf{k}_j}{\sqrt{d}},
    \quad 
    p_j = \frac{\exp(s_j)}{\sum_{k \in \mathcal{T}_c} \exp(s_k)},
    \quad
    \underbrace{\mathbf{k}'_c = \sum_{j \in \mathcal{T}_c} p_j \mathbf{k}_j}_{\mathrm{Attn}(\mathbf{q}'_c,\mathbf{K}_c,\mathbf{K}_c)},
    \quad 
    b'_c = 
    \underbrace{-\sum_{j\in \mathcal{T}_c} p_j \log p_j}_{\mathrm{Entropy}}.
    \label{eq:prop_q_c}
\end{equation}
\end{proposition}
Eq.~\eqref{eq:attn_lhsa} expresses the chunk-level surrogate score as the sum of a relevance term and a bias term. The learnable query $\mathbf{q}'_c$ serves as a proxy for queries that may attend to chunk $c$, inducing an attention distribution over its keys. 
The weighted sum of keys defines the chunk summary $\mathbf{k}'_c$, and $\mathbf{q}^{\top}\mathbf{k}'_c/\sqrt d$ measures its relevance to the current query. 
The bias $b'_c$ is the entropy of this distribution, capturing token-level mass beyond the linear relevance term. Specifically, $b'_c$ adaptively interpolates between two regimes in Eq~\eqref{eq:two_regime}: it approaches $\log S$ for nearly uniform scores and $0$ when a single score dominates. 
Both $\mathbf{k}'_c$ and $b'_c$ can be computed via $\mathrm{Attn}(\mathbf{q}'_c,\mathbf{K}_c,\mathbf{K}_c)$. 
The computational cost is $O(S)$ per chunk; therefore, for a sequence of length $N$ with $N/S$ chunks, the total cost is $O(N)$.

To instantiate $\mathbf{q}'_c$, we append a special \textit{landmark token}~\cite{mohtashami2023random} to the end of each chunk, and use its query vector as $\mathbf{q}'_c$. The resulting pair $(\mathbf{k}'_c,b'_c)$ serves as an entropy-calibrated compressed key for chunk-level routing: we score each chunk by the linear surrogate in Eq.~\eqref{eq:attn_lhsa}, select the top-$K$ chunks, and estimate the partition function as follows:
\begin{equation}
\label{eq:surrogate_topk}
\small
\begin{aligned}
    &\hat{s}_{i,c} = \frac{\mathbf{q}_i^\top \mathbf{k}'_c}{\sqrt{d}} + b'_{c}, \quad \bm{\mathcal{I}}_i=\left\{
    c \in \mathcal{C}_i
    \mid
    \operatorname{rank}_{\downarrow}(\hat{s}_{i,c}) < K
    \right\},\\
    &\hat{Z}_{i,c} = \exp(\hat{s}_{i,c}), \quad \hat{\mathcal{Z}}_i = \sum_{c \in {\mathcal{I}}_i} \hat{Z}_{i,c} + Z_{i,\text{swa}}, \\
\end{aligned}
\end{equation}
This enables a native sparse-training implementation: each query routes over all $N/S$ chunk summaries, keeps the top-$K$ chunks, and attends only to constant selected tokens. Thus, the routing cost, $O(N/S)$ per token and $O(N^2/S)$ per sequence, is the only quadratic term.

In conclusion, RQ1 is answered by showing that the LogSumExp chunk mass $\log Z_{i,c}$ can be expressed in the form of Eq.~\eqref{eq:attn_lhsa}, where each chunk is represented by a compact summary $(\mathbf{k}'_c,b'_c)$ derived from the learnable landmark query $\mathbf{q}'_c$.

\paragraph{Answering RQ2: Hierarchical softmax.} 
To make $\mathbf{q}'_c$ learnable, a core idea is to enable $\hat{Z}_{i,c}$ to participate in the forward pass of attention mass computation. Thus, we factorize the attention mass into an intra-chunk normalized term and an inter-chunk mass term, where the latter can be replaced by the learnable mass surrogate $\hat{Z}_{i,c}$. For tokens in selected chunks and the local sliding window, their normalized attention weight can be reformulated as
\begin{equation}
\label{eq:hils-attn-weights}
\small
w_{i,j}=\frac{\exp(s_{i,j})}{\mathcal{Z}_i}=\underbrace{\frac{\exp(s_{i,j})}{Z_{i,c(j)}}}_{\mathrm{intra-chunk}}\times \underbrace{\frac{Z_{i,c(j)}}{\mathcal{Z}_i}}_{\mathrm{inter-chunk}}\approx \frac{\exp(s_{i,j})}{Z_{i,c(j)}}\times \underbrace{\frac{\hat{Z}_{i,c(j)}}{\hat{\mathcal{Z}}_i}}_{\mathrm{surrogate}},
\end{equation}
where $\hat{Z}_{i,c(j)}=Z_{i,\mathrm{swa}}$ when $\ell(i)\le j \le i$. Intuitively, the model first aggregate relevant information within each selected chunk, then fuse chunk-level information according to the learned surrogate mass. Since $\hat{Z}_{i,c}$ impacts the final
attention weights, gradients from the LM loss can supervise the landmark representation learning and encourage chunks more useful for prediction to receive a larger mass. Empirically, this self-supervised surrogate outperforms naive BSA (Tab.~\ref{tb:345M_main_ppl} and Tab.~\ref{tb:345M_main_ruler}), suggesting that end-to-end learning
offers more effective allocation of attention mass.

\section{Practical Design Choices}

This section details our practical design choices on HiLS implementation across model architecture, GPU kernel design, and continuous training recipe.

\subsection{Architectural Design}

We follow mainstream open-source Transformer architectures such as Qwen3~\cite{yang2025qwen3} but replacing the standard full-attention with HiLS attention. In addition, we empirically find that slight architectural adjustments can better unleash the potential of HiLS attention, which we introduce below.
\paragraph{Positional Encoding.} 

Empirically, when training with an 8K context length, we find that HiLS attention performs worse than the full-attention baseline in perplexity with standard RoPE~\cite{su2024roformer}. Replacing RoPE with a partial RoPE variant, however, enables HiLS attention to outperform full attention in perplexity. Specifically, we adopt HoPE~\cite{chen-etal-2025-hope} positional encoding, which retains the RoPE dimensions whose rotation periods do not exceed the pre-training context length, and replaces the remaining dimensions with NoPE. 

\paragraph{Low-Rank Query Calibration.} 
In Eq.~\eqref{eq:surrogate_topk}, 
$\hat{\mathcal{Z}}_i = \sum_{c} \hat{Z}_{i,c} + Z_{i,\text{swa}}$ 
combines chunk-level mass surrogates with token-level attention mass. 
Since the chunk representation $\mathbf{k}'_c$ is not a token key but a compressed summary of multiple tokens, the original token-level query $\mathbf{q}_i$ may not be optimal for estimating chunk-level mass. 
To better calibrate chunk scores, we introduce a lightweight low-rank query calibration (Q-Cal) module for the chunk-level surrogate.
Empirically, this module significantly improves perplexity and length extrapolation.
Specifically, denoting the hidden state of $x_i$ as $\mathbf{h}_i \in \mathbb{R}^{d_{\text{model}}}$,
the adapted query $\hat{\mathbf{q}}_i$ and chunk score $\hat{s}_{i,j}$ are formulated as:
\begin{equation}
\label{eq:lora_q}
\small
    \Delta \mathbf{q}_i = \mathbf{W}^{\text{up}}\mathbf{W}^{\text{down}}\mathbf{h}_i, \quad \hat{\mathbf{q}}_i = \mathbf{q}_i + \Delta \mathbf{q}_i,\quad \hat{s}_{i,c} = \frac{\hat{\mathbf{q}}_i^\top \mathbf{k}'_c}{\sqrt{d}} + b'_{c}
\end{equation}
where $\mathbf{W}^{\text{up}} \in \mathbb{R}^{d \times r}$ and $\mathbf{W}^{\text{down}} \in \mathbb{R}^{r \times d_{\text{model}}}$ represent the low-rank projection weights with $r \ll d_{\text{model}}$.

\paragraph{HiLS-Attention in GQA.} 
Modern LLMs often adopt grouped-query attention (GQA)~\cite{ainslie-etal-2023-gqa} to reduce KV cache memory, 
where multiple query heads share the same key-value head. This requires adapting the chunk-selection procedure of HiLS-Attention from the standard MHA setting. In MHA, each query head can independently select its own top-$K$ chunks. In GQA, however, query heads within the same group should use the same retrieved chunk set, so that the selected tokens can be gathered once and processed efficiently with batched attention kernels. However, query heads in the same GQA group may attend to different chunks. To preserve head-level flexibility under the shared-retrieval constraint, we compute normalized chunk weights for each query head separately, aggregate them by taking the maximum over heads within the group, and select the top-$K$ chunks using these group-level scores. In this way, a chunk can be selected if it is important to any head in the group. We provide the full formalization of this GQA-aware chunk selection procedure in Appendix~\ref{appdx:hils-attn-gqa}.

\paragraph{Landmark token free alternative.} To avoid the engineering and implementation overhead of landmark tokens, we propose an alternative implementation. Instead of computing a unique query per chunk via landmark tokens, we employ a shared learnable query for each layer. Empirically, this approach achieves comparable in-domain performance, though its extrapolation capability degrades significantly. More details are given in the experiments.

\subsection{Hardware-Efficient Kernel Design}
An implementation challenge for sparse attention is that different tokens correspond to distinct chunk sets. Consequently, a naive implementation fails to achieve speedups and can cause memory explosion. In practice, efficient sparse attention requires hardware-software co-design.

\begin{figure}[htb!]
    \centering
    \begin{subfigure}[b]{0.48\textwidth}
        \centering
        \includegraphics[width=\linewidth]{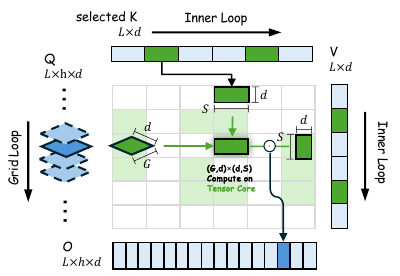}
        \caption{NSA kernel}
        \label{fig:kernel_design_NSA}
    \end{subfigure}
    \begin{subfigure}[b]{0.48\textwidth}
        \centering
        \includegraphics[width=\linewidth]{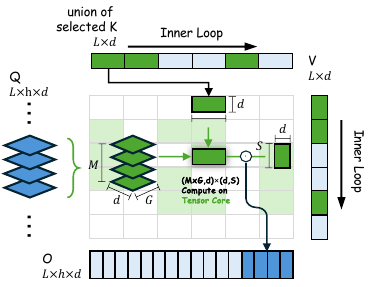}
        \caption{HiLS-Attention kernel}
        \label{fig:kernel_design_hils}
    \end{subfigure}
    \caption{Kernel design of NSA and HiLS-Attention. 
    Kernel design of NSA and HiLS-Attention.
    (a) NSA handles one query token per tile and computes attention over its selected chunks. Each Tensor Core operation has shape $(G,d)\times(d,S)$, where $G$ is the GQA group size and $S$ is the chunk size.
    (b) HiLS-Attention packs $M$ adjacent query tokens, attends to the union of their selected chunks, and enlarges the Tensor Core operation to $(M\times G,d)\times(d,S)$. This packing reuses overlapping K/V chunks across adjacent tokens.}
    \label{fig:kernel_design}
\end{figure}

NSA~\cite{yuan-etal-2025-native} provides a representative kernel design. As shown in Fig.~\ref{fig:kernel_design_NSA}, NSA processes one query token at a time and loads its selected K/V chunks. For each selected chunk, the Tensor Core computation has shape $(G,d)\times(d,S),$
where $G$ is the number of query heads sharing the same K/V head under GQA, $d$ is the head dimension, and $S$ is the chunk size. Since Tensor Cores are most efficient when the matrix tile dimension is at least $16$, MHA with $G=1$ leads to severe under-utilization, effectively padding the query-head dimension from $1$ to $16$. NSA mitigates this issue by relying on GQA with $G\geq16$. However, this implicitly requires a query-to-KV head ratio of at least $16$, which limits its applicability.

Instead, HiLS-Attention batches computation across both query tokens and query heads. Since adjacent query tokens often retrieve highly overlapping chunks, with the top-$k$ overlap ratio reported to be as high as $80\%$~\citep{huang2026nosanativeoffloadablesparse}, we group $M$ adjacent query tokens and compute attention over the union of their selected chunks, as illustrated in Fig.~\ref{fig:kernel_design_hils}. This changes the Tensor Core computation to $(M\times G,d)\times(d,S),$
so efficient Tensor Core utilization only requires $M\times G\geq16$, rather than $G\geq16$. Meanwhile, taking the union of selected chunks allows the packed queries to reuse loaded K/V chunks, reducing redundant memory access. This design does not require a large GQA group size and can also be applied to speculative decoding~\cite{leviathan2023fast}.

\subsection{Continuous Training Strategies}\label{sec:continue-training}
To preserve the model's original capabilities, we evaluate two continued-training strategies: (1) a lightweight, nearly training-free setup that freezes the base model and updates only the newly introduced parameters, and (2) full-parameter tuning.

\begin{itemize}[leftmargin=*,itemsep=0.2em,topsep=0.3em]
\item \textbf{Landmark Token Tuning:}
In this setting, all base model parameters are frozen. The only trainable parameters are the landmark token embeddings and the projection matrices $\mathbf{W}^{\text{up}}, \mathbf{W}^{\text{down}}$ in Eq.~\eqref{eq:hils-attn-weights}, accounting for less than 1\% of total parameters. Empirically, we find that training on no more than 5B tokens is sufficient to achieve performance comparable to the base model. 
\item \textbf{Full-Parameter Tuning:}
In this setting, the landmark token embeddings and low-rank query adapter parameters are randomly initialized, while all other parameters are inherited from the base model. All parameters are then jointly updated. This strategy is particularly effective when replacing the positional encoding with HoPE, thereby maximizing length generalization.
\end{itemize}

\section{Small-scale  Studies}
We first conduct small-scale experiments to rigorously evaluate HiLS-Attention against diverse baselines and investigate the empirical impact of each core component on its performance, in order to gain further insight on the behavior of HiLS-Attention.

\subsection{Main experiments}\label{sec:small_scale_exp}

\paragraph{Setup.}

Following the architectural configurations of GPT-2 Medium~\citep{radford2019language}, we train decoder-only transformers from scratch at the 345M parameter scale with different attention methods. All the models are initialized and trained using the same recipe with 8K context length (see details in Appendix~\ref{apdx:training_recipe}). For all the sparse attention variants, we allocate a 2K-token activation budget alongside a 512-token local sliding window, following findings of prior works that a 512-token window provides a strong efficiency-quality trade-off~\cite{wang2025rattentionminimalslidingwindow}.

RULER~\citep{hsieh2024ruler} is a synthetic long-context task that inserts specified needles into the context and asks corresponding questions at the end, requiring the model to retrieve the relevant needles from the context. It is therefore commonly used to evaluate a model's ability to retrieve task-relevant information from long inputs.
Because small models have limited instruction-following ability, standard RULER-style prompts may not 
provide a clean measurement of in-context retrieval. We therefore convert 5\% of the training samples into RULER-style Needle-in-a-Haystack (NIAH) tasks. This is achieved by inserting multiple needles into the context and appending the corresponding query and answer at the end. 

\paragraph{Baselines.}
We benchmark HiLS-Attention against three categories of attention methods: 
(1) \textit{Full Attention}, and its extrapolation-enhanced variant with HoPE position embedding \textit{Full-Attn (HoPE)}~\citep{chen-etal-2025-hope};
(2) \textit{Sliding-Window Attention (SWA)},   with a fixed 512-token window size across all layers; and
(3) \textit{State-of-the-Art Sparse Attention}, including Native Sparse Attention (NSA)~\citep{yuan-etal-2025-native}, DashAttention~\citep{huang2026dashattentiondifferentiableadaptivesparse}, InfLLM v2~\citep{zhao2025infllmv2densesparseswitchableattention} and HSA-UltraLong~\citep{hu2026everytoken}. For fair comparison, we use the same sparsity-related hyperparameters across all sparse baselines, including the local sliding-window size 512, chunk size 64, and top-$K$ 32. This ensures that different methods operate under a comparable attention budget.

\paragraph{Evaluation Metrics.}
As small models have limited instruction-following ability, we assess model capability via \emph{language modeling perplexity (PPL)} and \emph{in-context retrieval} performance. We quantify retrieval capability using the RULER benchmark~\citep{hsieh2024ruler}, focusing on three representative evaluation primitives: Single Needle (S-N), Multi-Key Multi-Query (MK-MQ, with 6 KV pairs and 2 ordered retrieval queries), and Variable Tracking (VT, comprising 6 assignments spanning up to 2 pointer hops). We argue that evaluating sparse attention on RULER should be conducted on small-scale models, since most tasks are essentially ``find \& copy'' and larger models may compensate for retrieval inaccuracies with broader information bandwidth. 

\newsavebox{\hiddenbox}
\newcolumntype{Y}{>{\begin{lrbox}{\hiddenbox}}c<{\end{lrbox}}}

\begin{table}[tb!]
\centering
\caption{Perplexity (PPL) of models with 345M parameters trained on 8K context and evaluated with different context lengths (from 64 to 512K). For fair comparison,  some baselines add the extra 0.6\% parameters introduced by HiLS-attention to their MLP modules, denoted as ``extra MLP''. Since the InfLLM v2 kernel only supports head dimensions $\geq 128$, it is not directly aligned with the other models using head dimension 64. We mark InfLLM v2 with an asterisk (*) to indicate that its results are reported for reference. \textbf{Bold} numbers indicate the best PPL at each context length, and \underline{underlined} numbers indicate the second-best PPL. Overall, HiLS-Attention achieves consistently strong performance across context lengths and demonstrates better long-context extrapolation.}
\label{tb:345M_main_ppl}
\vspace{0.1cm}
\resizebox{1.\textwidth}{!}{
\begin{tabular}{l@{\hspace{\tabcolsep}}Y@{\hspace{\tabcolsep}}cccccccc}
\toprule
Models & \#steps & Extra \#Param & 64 & 128 & 512 & \textbf{8K} & 32K & 128K & 512K \\
\midrule
Full-Attn RoPE & 30k & -- & 33.92 & 26.89 & 18.68 & 4.96 & $>10^2$ \\
Full-Attn HoPE            &     30k &  --  &  34.15 & 26.97 & 18.73 & \underline{4.95} &  6.42 & $>10^2$   \\
Full-Attn HoPE (extra MLP)        &     30k &  0.6\%  &  34.28 & 27.12 & 18.76  & \underline{4.95} &  5.85 & $>10^2$   \\
SWA-RoPE                 &     30k &  --  &   35.38 & 27.94 & 19.30 & 8.95 & 9.01 & 8.44 & 8.47 \\
NSA-RoPE   &   30k &  0.2\%  &  34.15 & 27.11 & 18.85 & 5.01 & 7.62 & 11.75 & 19.14 \\
Dash-Attention-RoPE   &   30k &  0.006\%  &   \underline{33.70} & \underline{26.74} & \underline{18.67} & 5.00 &  $>10^2$ \\
HSA-Ultralong         &     30k &   12.7\% &   \textbf{33.19} & \textbf{26.26} & 21.50 &  8.81 & 5.84 & \underline{4.99} &   \textbf{4.54} \\
Naive-BSA-HoPE & & -- & 36.12 & 29.06 & 20.45 & \textbf{4.94} & \textbf{3.94} & 8.67 & OOM \\
\textbf{HiLS-Attn-HoPE}        &     30k &  0.6\%  &  33.97 & 26.91 & \textbf{18.65}  & \textbf{4.94} &  \underline{4.34} & \textbf{4.71}  & \underline{5.95} \\
\textcolor{gray}{InfLLM v2$^*$}  &   \textcolor{gray}{30k} &  \textcolor{gray}{0.006\%}  &  \textcolor{gray}{33.71} & \textcolor{gray}{26.74} & \textcolor{gray}{18.55} & \textcolor{gray}{4.95} & \textcolor{gray}{13.68} & \textcolor{gray}{64.24} & \textcolor{gray}{OOM} \\
\bottomrule
\end{tabular}
}

\end{table}

\begin{table}[htb!]
\centering
\footnotesize
\caption{
RULER results for 345M models trained with an 8K context length. 
We additionally report ultra-long extrapolation results of HiLS-Attn-HoPE at 1M--4M context lengths. HiLS is the only native sparse attention method that achieves perfect in-domain NIAH performance, while also demonstrating remarkable extrapolation capability.
}  
\label{tb:345M_main_ruler}
\setlength{\tabcolsep}{1.pt} 
\vspace{2mm}

\begin{tabular}{l ccc >{\columncolor{blue!10}}c>{\columncolor{blue!10}}c>{\columncolor{blue!10}}c ccc >{\columncolor{blue!10}}c>{\columncolor{blue!10}}c>{\columncolor{blue!10}}c ccc}
\toprule
& \multicolumn{3}{c}{\textbf{8K}} & \multicolumn{3}{c}{\textbf{16K}} & \multicolumn{3}{c}{\textbf{32K}} & \multicolumn{3}{c}{\textbf{128K}} & \multicolumn{3}{c}{\textbf{512K}} \\
\cmidrule(lr){2-4} \cmidrule(lr){5-7} \cmidrule(lr){8-10} \cmidrule(lr){11-13} \cmidrule(lr){14-16}
\textbf{Models} & S-N & MK-MQ & VT & S-N & MK-MQ & VT & S-N & MK-MQ & VT & S-N & MK-MQ & VT & S-N & MK-MQ & VT \\
\midrule
Full-Attn RoPE             & \textbf{100} & 97 & 34 &   0 &  0 &  0 &  0 &  0 &  0 &  0 &  0 &  0 &  0 &  0 &  0 \\
Full-Attn HoPE             & \textbf{100} & \underline{98} & 36 &  \underline{99} & 97 & \underline{21} & 72 & 11 & 11 &  0 &  0 &  0 &  0 &  0 &  0 \\
Full-Attn HoPE (extra MLP) & \textbf{100} & \textbf{99} & \underline{39} & \textbf{100} & \textbf{97} & 13 & 83 & 19 &  1 &  0 &  0 &  0 &  0 &  0 &  0 \\
SWA-RoPE                   &   18 &  8 & 22 &   4 &  1 & 11 &  8 &  2 &  7 &  2 &  1 & 12 &  0 &  1 &  8 \\
NSA-RoPE                   &   49 & 17 & 16 &   6 &  5 &  4 &  5 &  0 &  0 &  0 &  0 &  0 &  0 &  0 &  0 \\
Dash-Attention-RoPE        &   \underline{92} & 67 & 34 &   0 &  0 &  0 &  0 &  0 &  0 & -- & -- & -- & -- & -- & -- \\
HSA-Ultralong              &   87 & 79 & 23 &  92 & 75 & 17 & 82 & 74 & \underline{15} & \underline{80} & \underline{60} & 21 & \underline{65} & \underline{42} &  \underline{8} \\
Naive-BSA-HoPE             & \textbf{100} & 97 & 23 &  \underline{99} & 89 & 16 & \underline{98} & \underline{77} & 13 & 46 &  0 &  0 & -- & -- & -- \\
\textbf{HiLS-Attn-HoPE}    & \textbf{100} & 95 & \textbf{72} & \textbf{100} & \underline{94} & \textbf{65} & \textbf{100} & \textbf{95} & \textbf{68} & \textbf{99} & \textbf{95} & \underline{61} & \textbf{99} & \textbf{91} & \textbf{66} \\
\textcolor{gray}{InfLLM v2*} &  \textcolor{gray}{77} & \textcolor{gray}{39} & \textcolor{gray}{25} &  \textcolor{gray}{21} &  \textcolor{gray}{0} &  \textcolor{gray}{1} &  \textcolor{gray}{1} &  \textcolor{gray}{0} &  \textcolor{gray}{0} &  \textcolor{gray}{0} &  \textcolor{gray}{0} &  \textcolor{gray}{0} & \textcolor{gray}{--} & \textcolor{gray}{--} & \textcolor{gray}{--} \\
\bottomrule
\end{tabular}

\vspace{2mm} 

\begin{tabular}{l ccc ccc ccc}
\multicolumn{10}{l}{\textit{Results on longer context length}} \\
\toprule
& \multicolumn{3}{c}{\textbf{1M}} & \multicolumn{3}{c}{\textbf{2M}} & \multicolumn{3}{c}{\textbf{4M}} \\
\cmidrule(lr){2-4} \cmidrule(lr){5-7} \cmidrule(lr){8-10}
\textbf{Models} & S-N & MK-MQ & VT & S-N & MK-MQ & VT & S-N & MK-MQ & VT \\
\midrule
HiLS-Attn-HoPE & 100 & 97 & 53 & 97 & 87 & 50 & 96 & 89 & 43 \\
\bottomrule
\end{tabular}
\end{table}

\paragraph{Results and Analysis.}
Tab.~\ref{tb:345M_main_ppl} and \ref{tb:345M_main_ruler} present the small-scale evaluation results. 
Based on these empirical results, we have the following key observations. 

\begin{itemize}[leftmargin=*,itemsep=0.2em,topsep=0.3em]
    \item \textbf{Expressiveness limitations prevent perfect NIAH, even under in-domain settings.} On the simplest Single-NIAH task, only Naive BSA and HiLS-Attention sustain performance comparable to full attention within the 8K in-domain context length. By contrast, existing sparse attention that use mean-pooled summaries, including NSA, DashAttention, and InfLLM v2, show noticeable degradation.
    This supports our analysis in Sec.~\ref{sec:exp_limit}: mean-pooled chunk summaries have limited expressiveness, as they dilute highly concentrated attention mass—precisely the pattern required by NIAH, where only a few needle tokens dominate.
    \item 
    \textbf{HiLS-Attention learns more robust sparse attention allocation than naive BSA.}
    HiLS-Attn achieves PPL comparable to naive-BSA at the training length (4.94 v.s. 4.94 at $8$K), but performs much better under context interpolation ($\leq 256K$) and extrapolation ($> 256K$).
    Together with the RULER results (where HiLS significantly leads for both within and beyond the training context length settings), this suggests that HiLS does not simply mimic the behavior of full-attention and estimate the full-attention-induced chunk mass, but it indeed learns more effective sparse attention allocation over important context tokens. A possible reason is that token-level full attention has an inherent noise issue: \textit{as long as a token logit is not $-\infty$, the token receives non-zero attention mass.} Since the model cannot learn $-\infty$ attention logits for all irrelevant tokens in practice, these irrelevant tokens inevitably consume small but nonzero probability mass, injecting noise to the probability mass. 
    As the context grows longer, these small noisy masses accumulate over many irrelevant tokens, making the full-attention-induced chunk mass a noisy selection signal. This may lead naive BSA to suboptimal chunk selection, which is consistent with its inferior downstream performance compared with HiLS.
    
    \item \textbf{Compression Enhances In-context Retrieval.}
    HiLS-Attn not only outperforms naive-BSA, but also substantially surpasses full attention on variable tracking (VT), a task that requires multi-hop reasoning. 
    We attribute this gain to compression: \textit{it allows token-level noise to partially cancel out while preserving shared semantic signals, thereby yielding more reliable retrieval representations.}
    Concretely, if each key can be decomposed as $\mathbf{k}_i=\mathrm{semantic}(\mathbf{k}_i)+\mathrm{noise}(\mathbf{k}_i)$, then compression aggregates multiple keys into $\mathbf{k}'_c$. The noise terms, being less aligned, tend to cancel out, while the shared semantic signal is preserved, yielding a cleaner representation. Therefore, compression may lead to better in-context retrieval. From this perspective, we conjecture that HiLS's strong extrapolation ability stems from its improved in-domain retrieval capability.

\end{itemize}

\subsection{Long-context Scaling}
Under the aforementioned setting, the sparsity of HiLS-Attention remains no less than 25\%, i.e., 2K out of an 8K context.
To evaluate the model's performance under even higher sparsity, we extend the training context length to 256K while keeping a maximum of 2K tokens activated, thereby validating its capabilities in both sparser and longer-context regimes.

\paragraph{Continued-Training Setup.} 
To validate HiLS-Attention under a longer training context length, we continue training the 345M checkpoints from the previous 8K-context models using a 256K context length.
Before continued training, we enlarge the RoPE base from $1\times 10^4$ to $1\times 10^7$ for the RoPE-based positional components to expand the perceptible range of global positions. The continued training data are from \texttt{allenai/dolma3\_longmino\_mix-50B-1025}~\citep{allenai_dolma3_longmino_mix_50b_1025}, i.e., the long-context mixture used for Olmo 3 7B long-context training. 
We train both the Full-Attn and HiLS-Attn-HoPE models under the same settings.

\begin{table}[tb]
\centering
\caption{Perplexity for 345M models continue trained with a 256K context length on 10B tokens.
}\label{tbl:345M_256K_ppl}
\vspace{0.1cm}
\resizebox{0.75\textwidth}{!}{
\begin{tabular}{lccccccc}
\toprule
Models & Extra \#Param &
8K & 32K & 128K & 256K & 512K & 1M \\
\midrule

\multicolumn{8}{l}{\footnotesize \textcolor{blue!60}{\textit{RoPE} $\theta = 1\times 10^7$} } \\
\hspace{1em} Full-Attn RoPE
& --
& 9.11
& 8.86
& 8.11
& 7.49
& 7.54
& 8.61 \\

\hspace{1em} \textbf{HiLS-Attn-HoPE}
& 0.6\%
& 9.08
& 8.88
& 8.15
& 7.45
& 7.37
& 8.08 \\
\midrule
\multicolumn{8}{l}{\footnotesize \textcolor{blue!60}{\textit{RoPE} $\theta = 1\times 10^4$ }} \\
\hspace{1em} Full-Attn RoPE 
& --
&9.22
&9.17
&8.54
&7.87
&7.82
&9.18
\\

\hspace{1em} Full-Attn HoPE 
& --
&9.20
&8.93
&8.15
&7.53
&$>10^2$
&
\\

\hspace{1em} \textbf{HiLS-Attn-HoPE} & 0.6\%
& 9.09 & 8.89 & 8.17 & 7.51 & 7.55 & 8.44 \\

\bottomrule
\end{tabular}
}
\end{table}

\begin{table}[tb]
\centering
\small
\caption{RULER results for 345M models continue trained with a 256K context length on 10B tokens.}\label{tbl:345M_256K_ruler}
\vspace{0.1cm}
\label{tb:ruler_eval_345M}
\setlength{\tabcolsep}{1.pt} 
\resizebox{1.0\textwidth}{!}{
\begin{tabular}{l ccc >{\columncolor{blue!10}}c>{\columncolor{blue!10}}c>{\columncolor{blue!10}}c ccc >{\columncolor{blue!10}}c>{\columncolor{blue!10}}c>{\columncolor{blue!10}}c ccc >{\columncolor{blue!10}}c>{\columncolor{blue!10}}c>{\columncolor{blue!10}}c ccc}
\toprule

& \multicolumn{3}{c}{\textbf{8K}}
& \multicolumn{3}{c}{\textbf{32K}}
& \multicolumn{3}{c}{\textbf{128K}}
& \multicolumn{3}{c}{\textbf{256K}}
& \multicolumn{3}{c}{\textbf{512K}}
& \multicolumn{3}{c}{\textbf{1M}} \\
\cmidrule(lr){2-4} \cmidrule(lr){5-7} \cmidrule(lr){8-10} \cmidrule(lr){11-13} \cmidrule(lr){14-16} \cmidrule(lr){17-19}
\textbf{Models}
& {\small S-N} & {\small MK-MQ} & {\small VT}
& {\small S-N} & {\small MK-MQ} & {\small VT}
& {\small S-N} & {\small MK-MQ} & {\small VT}
& {\small S-N} & {\small MK-MQ} & {\small VT}
& {\small S-N} & {\small MK-MQ} & {\small VT}
& {\small S-N} & {\small MK-MQ} & {\small VT} \\
\midrule
\multicolumn{8}{l}{\footnotesize \textcolor{blue!60}{\textit{RoPE} $\theta = 1\times 10^7$} }   \\
\hspace{1em} \small Full-Attn RoPE
& 100 & 2 & 1 
& 100 & 5 & 3  
& 100 & 5 & 7  
& 99 & 6 & 5  
& 41 & 1 & 0  
& 2 & 0 & 0  \\ 
\hspace{1em} \small HiLS-Attn-HoPE     & 100 & 2 & 51  
& 100 & 11 & 54  
& 99 & 6 & 50  
& 100 & 5 & 56 
& 97 & 13 & 51 
& 96 & 5 & 46 \\

\midrule
\multicolumn{8}{l}{\footnotesize \textcolor{blue!60}{\textit{RoPE} $\theta = 1\times 10^4$} }  \\
\hspace{1em} \small Full-Attn RoPE 
& 11 & 7 & 0
& 3 & 0 & 0
& 0 & 0 & 0
& 1 & 0 & 0
& 0 & 0 & 0
& 0 & 0 & 0
\\

\hspace{1em} \small Full-Attn HoPE
& 99 & 19 & 12
& 99 & 41 & 5
& 96 & 36 & 7
& 84 & 30 & 7
& 2 & 0 & 0
& 0 & 0 & 0
\\

\hspace{1em} \small HiLS-Attn-HoPE
& 100 & 87 & 42
& 98 & 87 & 44
& 100 & 75 & 54
& 98 & 79 & 42
& 97 & 64 & 19
& 96 & 64 & 22
\\

\bottomrule
\end{tabular}
}
\end{table} 

\paragraph{Results.}
After extending the training context to 256K, Tab.~\ref{tbl:345M_256K_ppl} shows that HiLS-Attention consistently achieves lower perplexity across all settings when evaluated at a length of 256K than full attention.

We observe several interesting phenomena. First, without enlarging the RoPE base, Full-Attn RoPE almost fails on all RULER tasks, showing that the original positional range is insufficient for 256K contexts. However, when the RoPE base is enlarged, both are near zero on MK-MQ, suggesting that changing the RoPE base breaks the retrieval pattern learned within 8K.
Surprisingly, HiLS-Attn-HoPE still performs better than Full-Attn RoPE on variable tracking (VT). 
This aligns with our earlier observation that compression can enhance in-context retrieval in \S~\ref{sec:small_scale_exp}. 
VT requires retrieving a small number of relevant variable assignments from a long context; under full attention, irrelevant tokens still receive small but nonzero attention mass, and their accumulated noise can dilute the useful retrieval signal. 
By contrast, HiLS uses compressed-key retrieval, where token-level noise can partially cancel out while shared semantic signals are preserved, leading to cleaner retrieval and better VT performance.

Second, HiLS-Attn-HoPE consistently outperforms Full-Attn HoPE/RoPE under the same 256K training setting, despite activating at most 2K tokens. This shows that HiLS-Attention's sparsity does not sacrifice model capability; instead, accurate sparse retrieval can yield lower perplexity and much stronger long-context retrieval ability.

Overall, these results suggest that HiLS-Attn provides a promising path toward ultra-long or even infinite-context training. 
Infinite-context training must rely on native sparse attention to keep the attention cost bounded, whose key challenge is to ensure that the sparsely selected blocks contain relevant KV states. 
The strong length-extrapolation ability of HiLS-Attn provides a highly feasible approach to this problem: it can learn retrieval patterns in short contexts and generalize them to much longer contexts, enabling reliable block selection with ultra-long training length. 
Combined with the fixed computation cost of sparse attention, this makes infinite-context training practically possible.

\subsection{Ablation Study}

\paragraph{ Configurations.}
To isolate the contributions of each component in HiLS-Attention, we benchmark the default setting against the variants 
in Tab.~\ref{tab:ablation-configuration}.

\begin{table}[t]
    \centering
    \small
    \renewcommand{\arraystretch}{1.5}
    \caption{Summary of ablation variants and their configuration details. }
    \resizebox{\linewidth}{!}{
    \begin{tabular}{p{0.2\linewidth} p{0.8\linewidth}}
    \toprule
    \textbf{Ablation Variants} & \textbf{Description} \\
    \midrule
    \multicolumn{2}{l}{\textit{Query Calibration (Q-Cal) Variants}} \\
    \cmidrule(lr){1-2}
    w/ Q-Cal ($r$) & Employ the low-rank query projection   in Eq.~\eqref{eq:lora_q} with rank $r$. \\
    \textcolor{RoyalBlue}{w/ full-rank} & \textcolor{RoyalBlue}{Replaces the low-rank projection with a full-rank linear transformation $\hat{\mathbf{q}}_i = \mathbf{W}^{\hat{Q}}\mathbf{h}_i$, where $\mathbf{W}^{\hat{Q}} \in \mathbb{R}^{d_{\text{model}} \times d_{\text{model}}}$.} \\
    w/o Q-Cal & Remove the extra $\Delta\mathbf{q}$ term from the query. \\
    \midrule
    \multicolumn{2}{l}{\textit{Chunk Summarization Variants}} \\
    \cmidrule(lr){1-2}
    w/o Prop.~\ref{prop:logsumexp} & Use the raw landmark token key directly as the routing key $\mathbf{k}'_c$ without Taylor expansion rectification in Eq~\eqref{eq:attn_lhsa}. \\
    \textcolor{RoyalBlue}{LMK-Attn} & \textcolor{RoyalBlue}{Align with Landmark Attention~\cite{mohtashami2023random} by removing the extra $\Delta\mathbf{q}$ and using the landmark key.} \\
    w/ mean pooling & Use mean-pooled keys as the summary $\mathbf{k}'_c$ for each chunk. \\
    \textcolor{RoyalBlue}{w/o lmk, shared $\mathbf{q}_c$} 
    & \textcolor{RoyalBlue}{Remove landmark tokens and replace $\mathbf{q}'_c$ in Eq.~\eqref{eq:prop_q_c} with a shared   query $\mathbf{q}_c$.} \\
    \bottomrule
    \end{tabular}
    }
    \label{tab:ablation-configuration}
\end{table}

\paragraph{Main Findings.} The ablation results in Tab.~\ref{tb:345M_ablation_ppl} and \ref{tb:345M_ablation_ruler} show that HoPE positional encoding, query augmentation, and landmark tokens all contribute to the overall performance improvement.

\newcolumntype{Y}{>{\begin{lrbox}{\hiddenbox}}c<{\end{lrbox}}}

\begin{table}[tb]
\centering
\caption{Ablation perplexity (PPL) of 345M models trained on 8K context.}\label{tb:345M_ablation_ppl} 
\vspace{0.1cm}
\resizebox{0.9\textwidth}{!}{
\begin{tabular}{l@{\hspace{\tabcolsep}}Y@{\hspace{\tabcolsep}}cccccccc}
\toprule
Models & \#steps & Extra \#Param & 64 & 128 & 512 & \textbf{8K} & 32K & 128K & 512K \\
\midrule
HiLS-Attn-HoPE \\
\midrule
~~~w/ Q-Cal ($r=64$)        &     30k &  0.6\%  &  33.97 & 26.91 & \underline{18.65} & \textbf{4.94} & 4.34 & \underline{4.71} & \underline{5.95} \\
~~~w/ Q-Cal ($r=128$)        &     30k &  1.2\%  &  \underline{33.89} & \textbf{26.77} & \textbf{18.61} & \underline{4.95} & \textbf{4.26} & \textbf{4.54} & \textbf{5.85}  \\
~~~w/ full-rank &     30k &  4.9\%  &  34.02 & 26.93 & 18.66 & \textbf{4.94} & 4.62 & 5.03 & 6.52 \\
~~~w/o Q-Cal        &     30k &  0\%  &  34.04 & \underline{26.86}  & 18.68 & 4.97 & 7.21 & 12.40 & 16.93 \\
~~~w/o Prop.~\ref{prop:logsumexp}        &     30k &  0.6\%  &  34.13 & 27.05 & 18.74 & 4.97 & \underline{4.28} & 4.73 & 6.61 \\
~~~LMK-Attn        &     30k &  0\%  & 34.02 & 26.93 & 18.75 & 4.98 & 6.36 & 10.97 & 14.10 \\
~~~w/ mean pooling       &     30k &  0.6\%  & \textbf{33.82}  & 27.14 & 19.03 & 5.04 & 4.85 & 5.55 & 8.00 \\

~~~w/o lmk, shared  $\mathbf{q}_c$           &     30k &   0.6\%  &  34.02 & 26.92 & \underline{18.65} & \textbf{4.94} & 4.71 & 5.50 & 7.94 \\
\midrule
HiLS-Attn-RoPE \\
\midrule
~~~w/ Q-Cal ($r=64$)         &     30k &   0.6\% &  34.61 &  27.30&  18.89&  5.00&  $>10^2$ \\
~~~w/ full-rank &     30k &  4.9\%  & 34.04 & 26.96 & 18.73 & 4.97 & $>10^2$ \\
~~~w/o Q-Cal         &     30k &  0\%  &  34.10 & 26.97 & 18.76 & 5.03 & 9.66 & 10.46 & 24.94 \\

\midrule
HiLS-Attn-NoPE \\
\midrule
~~~w/ Q-Cal ($r=64$)         &     30k &  0.6\%  & 36.57 & 28.94 & 20.01 & 5.13 & 9.69 & 37.46 & 89.89 \\
~~~w/ full-rank &     30k &  4.9\%  & 36.07 & 28.63 & 19.83 & 5.10 & 5.37 & 28.26 & 82.31 \\
~~~w/o Q-Cal         &     30k &  0\%  &  36.78 & 29.10 & 20.09 & 5.17 & 29.02 & $>10^2$ \\
\bottomrule
\end{tabular}
}

\end{table}

\begin{table}[tb]
\centering
\caption{
Ablation RULER results for 345M models trained with an 8K context length.}
\vspace{0.1cm}
\label{tb:345M_ablation_ruler}
\setlength{\tabcolsep}{1.pt} 

\resizebox{0.9\textwidth}{!}{%
\begin{tabular}{l ccc >{\columncolor{blue!10}}c>{\columncolor{blue!10}}c>{\columncolor{blue!10}}c ccc >{\columncolor{blue!10}}c>{\columncolor{blue!10}}c>{\columncolor{blue!10}}c ccc}
\toprule
& \multicolumn{3}{c}{\textbf{8K}}
& \multicolumn{3}{c}{\textbf{16K}}
& \multicolumn{3}{c}{\textbf{32K}}
& \multicolumn{3}{c}{\textbf{128K}}
& \multicolumn{3}{c}{\textbf{512K}}
\\
\cmidrule(lr){2-4} \cmidrule(lr){5-7} \cmidrule(lr){8-10} \cmidrule(lr){11-13} \cmidrule(lr){14-16}

\textbf{Models}
& {\small S-N} & {\small MK-MQ} & {\small VT}
& {\small S-N} & {\small MK-MQ} & {\small VT}
& {\small S-N} & {\small MK-MQ} & {\small VT}
& {\small S-N} & {\small MK-MQ} & {\small VT}
& {\small S-N} & {\small MK-MQ} & {\small VT}
\\
\midrule
\multicolumn{16}{l}{\textbf{HiLS-Attn-HoPE}} \\
\RulerRowHideLengths{~~~w/ Q-Cal ($r=64$)}{30k}{
\textbf{100},\underline{95},\underline{72},18,12;
\textbf{100},\textbf{94},\textbf{65},22,9;
\textbf{100},\textbf{95},\textbf{68},\underline{24},12;
\textbf{100},94,65,24,11;
\textbf{99},\textbf{95},\underline{61},\underline{22},16;
\textbf{100},97,68,20,15;
\textbf{99},\textbf{91},\textbf{66},\underline{16},13
}

\RulerRowHideLengths{~~~w/ Q-Cal ($r=128$)}{30k}{
\textbf{100},94,64,22,9;
\textbf{100},\underline{89},47,11,5;
97,\underline{92},\underline{64},22,17;
95,94,57,18,7;
95,89,57,\textbf{27},13;
99,93,56,13,12;
91,\underline{86},\underline{56},\textbf{21},10
}

\RulerRowHideLengths{~~~w/ full-rank}{30k}{
\textbf{100},\textbf{98},\textbf{75},\textbf{27},8;
96,88,57,12,9;
97,89,\underline{64},17,12;
96,97,62,9,12;
92,\underline{91},54,18,12;
91,91,66,15,13;
\underline{92},77,\underline{56},14,11
}

\RulerRowHideLengths{~~~w/o Q-Cal}{30k}{
\underline{99},77,49,15,9;
\underline{98},74,20,\underline{24},10;
88,64,23,11,13;
76,52,7,14,12;
71,33,5,3,8;
59,12,1,0,8;
39,1,0,3,5
}

\RulerRowHideLengths{~~~w/o Prop.~\ref{prop:logsumexp}}{30k}{
\textbf{100},93,69,20,9;
\textbf{100},\underline{89},\underline{60},19,12;
\underline{98},91,57,14,8;
97,89,63,19,11;
\underline{97},\underline{91},\textbf{63},\underline{22},8;
92,90,67,12,10;
83,83,52,6,10
}

\RulerRowHideLengths{~~~LMK-Attn}{30k}{
96,79,38,14,7;
76,47,23,8,14;
65,31,9,12,15;
48,26,5,10,11;
17,9,8,9,12;
14,3,1,3,12;
3,1,0,4,12
}

\RulerRowHideLengths{~~~w/ mean pooling}{30k}{
92,68,29,20,11;
82,34,11,12,10;
70,23,12,11,6;
72,7,1,12,11;
55,3,1,0,7;
44,5,1,0,9;
24,2,1,1,9
}

\RulerRowHideLengths{~~~w/o lmk, shared  $\mathbf{q}_c$ }{30k}{99,80,35,18,8;98,82,22,14,6;95,71,20,11,19;89,46,18,9,10;85,32,10,6,13;65,18,9,6,10;54,2,4,0,9}

\midrule
\multicolumn{16}{l}{\textbf{HiLS-Attn-RoPE}} \\
\RulerRowHideLengths{~~~w/ Q-Cal ($r=64$)}{30k}{
\underline{99},81,52,16,15;
0,0,0,0,0;
0,0,0,0,0;
0,0,0,0,0;
0,0,0,0,0;
0,0,0,0,0;
0,0,0,0,0
}

\RulerRowHideLengths{~~~w/ full-rank}{30k}{\textbf{100},90,57,16,13;
0,0,0,0,7;
0,0,0,0,1;
0,0,0,0,4;
0,0,0,0,0;
0,0,0,0,9;
0,0,0,0,10}

\RulerRowHideLengths{~~~w/o Q-Cal}{30k}{
84,74,29,22,11;
39,20,1,10,6;
1,0,0,2,7;
0,1,0,0,10;
0,0,0,0,10;
0,0,0,0,9;
0,0,0,0,4
}

\midrule
\multicolumn{16}{l}{\textbf{HiLS-Attn-NoPE}} \\
\RulerRowHideLengths{~~~w/ Q-Cal ($r=64$)}{30k}{
\textbf{100},90,64,15,10;
\underline{98},81,47,19,8;
68,41,21,15,13;
45,20,11,4,6;
22,15,4,7,7;
6,1,2,1,5;
7,2,0,3,5
}

\RulerRowHideLengths{~~~w/ full-rank}{30k}{
90,88,70,18,9;
85,73,53,23,12;
70,47,27,14,10;
45,32,23,10,8;
24,23,7,4,3;
16,11,2,4,5;
3,4,0,3,5
}

\RulerRowHideLengths{~~~w/o Q-Cal}{30k}{
96,41,26,3,8;
91,34,17,8,18;
32,4,2,4,10;
3,1,0,0,9;
1,0,0,0,5;
0,0,0,0,5;
0,0,0,0,10
}

\bottomrule
\end{tabular}
}

\end{table}

\begin{itemize}[leftmargin=*,itemsep=0.2em,topsep=0.3em]

    \item \textbf{Impact of Low-Rank Query Calibration:} Excluding the low-rank path (\textit{w/o Q-Cal} in Tab.~\ref{tb:345M_ablation_ppl}) severely degrades performance, whereas overly expanding the rank dimension (r=128 in Table~\ref{tb:345M_ablation_ruler}) conversely impairs length extrapolation. 
    We conjecture that $\Delta Q$ helps decouple token-level attention from the chunk-level mass surrogate, as they operate at different information level. Although this phenomenon is consistently observed in our experiments, the underlying mechanism is not yet fully understood, and we leave a more principled investigation to future work.
    
    \item \textbf{Necessity of Landmark Tokens:} While the landmark-free variant (\textit{w/o lmk}, shared $\mathbf{q}_c$) offers a viable in-domain alternative, its context extrapolation capability degrades rapidly. This demonstrates that specific landmark tokens are indispensable for robust generalization over ultra-long sequences. 
    As discussed in Sec.~\ref{sec:method}, the effectiveness of a chunk compressed-key is determined by the quality of its learned query $\mathbf{q}'_c$. Since the shared $\mathbf{q}_c$ is only a fixed set of parameters, its expressive power is inherently limited and may fail to capture diverse chunk-level semantics. By contrast, landmark-token queries are produced by the full Transformer computation, including both attention and MLP layers, allowing them to exploit the model's full capacity and adaptively encode each chunk. This explains why landmark tokens are essential for robust long-context extrapolation.

    \item \textbf{Efficacy of Positional Encodings:} Incorporating HoPE yields the best performance, significantly outperforming alternative methods such as RoPE and NoPE. 
    By applying RoPE only to dimensions whose rotation periods are covered during training and NoPE otherwise, HoPE avoids out-of-distribution positional rotations beyond the training length.
    Although HoPE brings marginal gains in length extrapolation and perplexity for full attention, its integration with HiLS-Attention delivers substantial gains in both in-domain performance and long-context extrapolation.
    We hypothesize that HoPE improves HiLS-Attention by reducing the interference of rotary positional encodings during chunk compression. Since chunk compression performs a weighted aggregation over keys $\mathbf{k}_i$, applying RoPE to all dimensions also aggregates different positional rotations, which can distort the resulting semantic representation. The NoPE dimensions in HoPE provide a position-independent semantic subspace, helping preserve chunk semantics and thereby improving both in-domain performance and length extrapolation.
\end{itemize}

\section{Large-scale Experiments}

\subsection{Training from scratch}
To further test whether native sparse training can match full attention at a larger scale, we train a 1.4B-parameter model from scratch on 300B tokens. We track perplexity and length extrapolation performance throughout training, and evaluate the final model on multiple downstream tasks. We use full attention with RoPE as the baseline, as it achieves lower perplexity than full attention with HoPE within 512-token context in Tab.~\ref{tb:345M_main_ppl}, forming a stronger baseline for benchmarks whose average sequence length is below 64 tokens.
We evaluate PPL and NIAH tasks every 20K steps to track extrapolation stability and their performance discrepancy during training. Details about the model hyperparameters and training recipe are given in Appendix~\ref{appdx:hyper-params}. Benchmark details are listed in Appendix~\ref{apdx:benchmarks}.

\begin{figure}[t]
    \centering
    \includegraphics[width=\linewidth]{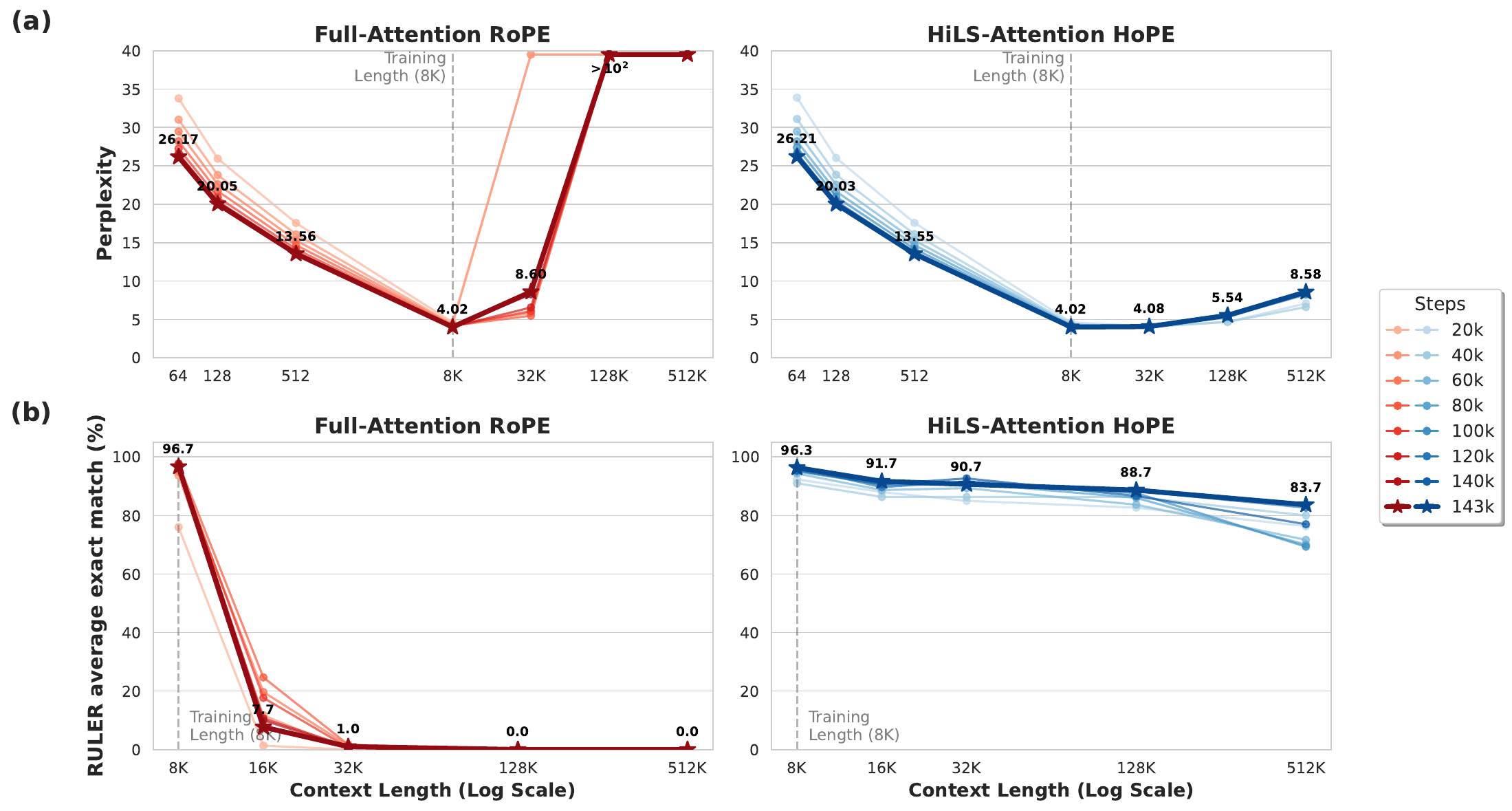}
    \caption{Perplexity (a) and RULER accuracy (b) of the 1.4B model at different training steps. Left: Full-Attention with RoPE;
    right: HiLS-Attention with  HoPE.
    The annotated values on the curves correspond to the final checkpoint (143k steps), which is highlighted with star markers and thicker lines. The detailed per-step results are deferred to Appendix~\ref{apdx:1B_perstep} (Tab.~\ref{tb:1B_ppl_steps} \& Tab.~\ref{tb:1B_ruler_steps}).}
    \label{fig:1B_perstep}
\end{figure}

\paragraph{Results.}
Fig.~\ref{fig:1B_perstep}(a) shows that the results are consistent with our small-scale findings: HiLS-attention and full-attention achieve almost identical PPL across different context lengths and training stages, especially close at 8K. This indicates that HiLS-attention, despite being trained with native sparsity, preserves short-context modeling ability that is fully comparable to full-attention. 
Figure~\ref{fig:1B_perstep}(b) confirms that HiLS-Attention's extrapolation remains stable without decaying over training.  Furthermore, Tab.~\ref{tb:1B_downstream} reveals that despite native sparse training\textemdash where HiLS-attention attends to less than half of the tokens of full-attention\textemdash  it still achieves comparable or better performance, thanks to its accurate retrieval. Overall, although HiLS-Attention modifies the standard attention form and is trained from scratch with native sparsity, it remains comparable to full-attention on short-context tasks while showing substantially stronger length extrapolation.

\begin{table}[tb]
\caption{Downstream task evaluation results for 1.4B model after 300B-token training.}\label{tb:1B_downstream}
\centering
\setlength{\tabcolsep}{3pt}
\vspace{0.1cm}
\resizebox{1.0\textwidth}{!}{
\begin{tabular}{lccccccccc >{\columncolor{blue!10}}c}
\toprule
\multirow{2}{*}{Models} &
LAMBADA & HellaSwag & PIQA & WinoGrande &
OpenBookQA & ARC-e & ARC-c &
ARC-e & ARC-c & AVG \\
 & &  &  &  & & 25-shot & 25-shot & &  &  \\
\midrule

Random-Guess
& -
& 25.00
& 50.00
& 50.00
& 25.00
& 25.00
& 25.00
& 25.00
& 25.00
& 25.00 \\

Full-Attn RoPE
& \textbf{57.13}
& 51.99
& 70.51
& \textbf{56.51}
& \textbf{22.60}
& 68.25
& 36.95
& 44.44
& \textbf{29.49}
& 48.65 \\

HiLS-Attn HoPE
& 57.05
& \textbf{52.21}
& \textbf{72.14}
& 55.64
& 22.50
& \textbf{70.02}
& \textbf{37.97}
& \textbf{45.86}
& 28.14
& \textbf{49.06} \\
\bottomrule
\end{tabular}
}
\label{tab:1b_downstream}
\end{table}

\subsection{Continue Pre-Training with HiLS-Attention} 
Since prompt lengths in most benchmarks rarely exceed 1,000 tokens — with Tab.~\ref{tb:1B_downstream} averaging under 64 tokens — they often fall entirely within HiLS-Attention’s sliding-window range. Therefore, we scale up the model size and evaluate on practical long-context tasks like LongBench~\cite{bai2024longbench} to validate its efficacy in real-world long-range scenarios.
\paragraph{Setup.} 
We start from the Olmo3-1025-7B base checkpoint~\citep{allenai_olmo3_1025_7b_checkpoint}, which uses standard MHA and a repeating pattern of three 4K sliding-window layers followed by one full-attention layer~\citep{olmo2025olmo}. 
 We replace the full-attention layers with HiLS-Attn and reduce the sliding-window size from 4K to 512 tokens, preserving the 3:1 pattern while shifting long-range modeling to the HiLS retrieval branch.

Following the small- and medium-scale settings, we set the chunk size to 64 and the HiLS top-$k$ to 32, which retrieves $32 \times 64 = 2048$ tokens under the 8K continue-pretraining length. Detailed training recipe can be found in Appendix~\ref{apdx:training_recipe}.

We compare HiLS-Attn variants against two baselines in Tab.~\ref{tab:olmo3_indomain}.

\textbf{LMK token tuning} follows the training-free recipe in Sec.~\ref{sec:continue-training}: all base parameters are frozen, and only the inserted landmark-token embeddings and the $\mathbf{W}^{\mathrm{up}}, \mathbf{W}^{\mathrm{down}}$ projections are updated.
In this stage, we conduct training on 5B tokens and use the same learning-rate schedule as above.

\textbf{Olmo3-512swa-CPT} is a full-parameter continued-pretraining baseline that keeps the original full-attention layers but reduces the sliding-window size in every SWA layer from 4K to 512 tokens, matching the local window used by our HiLS-Attn models.
This baseline isolates the effect of shrinking the local attention window without introducing HiLS retrieval; it uses the same 50B-token continued-pretraining recipe and optimizer schedule as the HiLS models.

\begin{table}[t!]
        \caption{
            General Downstream task results for OLMo3-7B continue pretraining.
        }
        \centering
        \small
        \setlength{\tabcolsep}{2pt}{
        \begin{tabular}{lcc|cccc}
        \toprule
        & \multicolumn{2}{c}{\textbf{Freezing param.}} & \multicolumn{4}{c}{\textbf{Full param. CPT}} \\
        \cmidrule(lr){2-3} \cmidrule(lr){4-7}
         \textbf{Benchmarks} &  \makecell{Olmo3\\Base} & \makecell{LMK token \\tuning} & \makecell{Olmo3 \\512swa} & \makecell{HiLS-Attn\\HoPE-Q-Cal} & \makecell{HiLS-Attn\\RoPE-Q-Cal} & \makecell{HiLS-Attn\\NoPE-Q-Cal}\\
        \midrule
        \textbf{Long-Context Retrieval} &   &  &  &  &  &    \\
        RULER-8K           & 11.34 & 22.33 & \textbf{99.67} & 99.00 & 98.67 & \textbf{99.67} \\
        RULER-16K           & 3.67 & 2.00 & 33.00 & 98.67 & 50.67 & 73.33 \\
        RULER-64K          & 0.00 & 1.00 & 1.33 & \textbf{97.33} & 4.33 & 0.33 \\
        RULER-128K         & 0.00 & 0.67 & 0.00 & \textbf{94.67} & 1.00 & 0.00 \\
        \rowcolor{blue!10}
        \textbf{Average}   & 3.75 & 6.50 & 33.50 & \textbf{97.42} & 38.67 & 43.33 \\
        \cmidrule(lr){1-7}
        \textbf{General Knowledge} &  &  &  &  &  &    \\
        MMLU(5-shot)       & \textbf{59.90} & 58.07 & \textbf{59.12} & 56.58 & 56.69 & 56.51 \\
        GPQA(5-shot)       & 29.29 & \textbf{32.32} & 31.31 & \textbf{34.34} & 24.75 & 29.80 \\
        Hellaswag(10-shot) & \textbf{44.17} & 43.25 & \textbf{42.96} & 38.71 & 33.17 & 34.24 \\
        ARC-c(25-shot)     & \textbf{53.56} & 52.88 & 55.59 & \textbf{55.93} & 54.92 & 53.90 \\
        BoolQ(5-shot)      & 61.01 & \textbf{61.10} & 64.22 & \textbf{64.71} & 63.43 & 63.43 \\
        Race(3-shot)       & \textbf{73.89} & 70.34 & \textbf{72.97} & 69.75 & 69.50 & 70.21 \\
        \cmidrule(lr){1-7}
        \textbf{Mathematics} &  &  &  &  &  &    \\
        CMath              & 41.53 & \textbf{42.08} & 39.98 & \textbf{43.35} & 42.44 & 40.16 \\
        GSM8K              & 37.00 & 37.00 & 33.43 & \textbf{36.85} & 35.71 & 35.03 \\
        \cmidrule(lr){1-7}
        \textbf{Code} &  &  &  &  &  &    \\
        CRUX               & 24.62 & 24.62 & 24.50 & 25.12 & 25.62 & \textbf{25.75} \\
        HumanEval+         & \textbf{20.10} & 19.50 & 19.50 & 18.90 & 18.90 & 17.70 \\
        MBPP+              & \textbf{37.60} & 33.80 & 32.30 & 32.60 & 33.30 & 31.10 \\
        \rowcolor{blue!10}
        \textbf{Average}   & \textbf{43.88} & 43.18 & 43.24 & \textbf{43.35} & 41.68 & 41.62 \\
        \bottomrule
        \end{tabular} }
        \label{tab:olmo3_indomain}
\end{table}

\begin{table}[t!]
    \centering
    \caption{Validation perplexity on the Olmo3 pretraining corpus at different context lengths. Lower is better. Models are continued-pretrained at 8K length. Perplexity values are reported to three decimal places to better highlight small differences across models.}
    \resizebox{0.8\textwidth}{!}{
    \begin{tabular}{lcccccc}
    \toprule
    Models & 512 & 8K & 16K & 64K & 128K & 256K \\
    \midrule
    Olmo3-base & 10.396 & 3.997 & 6.898 & 11.003 & 11.226 & 14.554 \\
    LDM token tuning & 10.394 & 3.470 & 5.124 & 7.058 & 6.126 & 7.770 \\
    Olmo3-512swa-CPT & \textbf{10.180} & \textbf{3.234} & 4.398 & 5.218 & 6.984 & 12.760 \\
    HiLS-Attn-HoPE & 10.201 & 3.236 & \textbf{2.772} & \textbf{2.722} & \textbf{2.550} & \textbf{3.095} \\
    HiLS-Attn-RoPE & 10.200 & 3.242 & 3.770 & 4.454 & 4.709 & 6.546 \\
    HiLS-Attn-NoPE & 10.235 & 3.248 & 2.895 & 27.117 & 51.911 & 73.896 \\
    \bottomrule
    \end{tabular}
    }

    \label{tab:olmo3_ppl}
\end{table}

\begin{table}[t!]
    \centering
    \small
    \caption{LongBench-v1 scores grouped by context length. Overall is weighted by the number of samples in each dataset.}
    \setlength{\tabcolsep}{1pt} 
    \vspace{0.1cm}
    \resizebox{\textwidth}{!}{
    \begin{tabular}{l  cccccc >{\columncolor{blue!10}}c>{\columncolor{blue!10}}c>{\columncolor{blue!10}}c>{\columncolor{blue!10}}c>{\columncolor{blue!10}}c>{\columncolor{blue!10}}c   c}
    \toprule
    & \multicolumn{6}{c}{\textbf{LongBench-v1 $<$8K}} & \multicolumn{6}{c}{\textbf{LongBench-v1 $>$8K}} & \textbf{Overall} \\
    \cmidrule(lr){2-7} \cmidrule(lr){8-13}
    \textbf{Method} & SDoc & MDoc & Summ. & Few-shot & Synth. & Code & SDoc & MDoc & Summ. & Few-shot & Synth. & Code & $\uparrow$ (\%) \\
    \cmidrule(lr){1-14}
    Olmo3-Base & 36.9 & 33.9 & 23.4 & 63.8 & \textbf{5.8} & 61.8 & 11.4 & 14.4 & 12.1 & 36.0 & 3.3 & 41.0 & 29.0 \\
    LDM token tuning & 32.1 & 30.4 & 20.6 & 62.2 & 4.1 & 60.9 & 14.3 & 17.5 & 13.3 & 43.1 & 2.6 & 49.2 & 28.8 \\
    Olmo3-512swa-CPT & 37.4 & 27.8 & 22.1 & 64.1 & 4.1 & 59.4 & 10.0 & 14.2 & 12.9 & 33.6 & 3.2 & 32.9 & 28.0 \\
     \; + YaRN 32K & 34.1 & 30.6 & 21.5 & 64.0 & 3.5 & 60.0 & 18.6 & 24.8 & 18.0 & \textbf{51.9} & 2.5 & 49.6 & 31.7 \\
    HiLS-Attn-HoPE & 36.4 & \textbf{37.8} & \textbf{24.6} & \textbf{64.6} & 4.7 & 59.3 & \textbf{23.3} & \textbf{25.0} & \textbf{18.1} & 50.7 & \textbf{5.3} & \textbf{51.0} & \textbf{33.2} \\
    HiLS-Attn-RoPE & 37.2 & 33.1 & 22.2 & 61.0 & 5.4 & 60.6 & 17.5 & 17.3 & 14.0 & 42.0 & 4.3 & 47.1 & 30.0 \\
    HiLS-Attn-NoPE & \textbf{37.5} & 34.3 & 23.6 & 64.4 & 5.4 & \textbf{62.3} & 22.3 & 24.9 & 17.8 & 50.9 & 4.2 & 48.7 & 33.2 \\
    \bottomrule
    \end{tabular}
    }
    \label{tab:longbench_v1}
\end{table}

\paragraph{Results.} 
The results on the 7B model are generally consistent with those on small- and medium-scale models: HiLS results are overall comparable or even slightly better than baselines (including full-attention) on general tasks.
We note that the slightly higher 8K perplexity of HiLS-Attn-HoPE-Q-Cal in Tab.~\ref{tab:olmo3_ppl} (3.234 v.s. 3.236) mainly comes from the migration cost of continuing training from a full-attention checkpoint, which is not observed from native HiLS-Attention training from scratch. The gap consistently shrinks with more training tokens, and detailed results are provided in the Appendix~\ref{apdx:ppl_in_cpt}.
Considering that most tasks in Tab.~\ref{tab:olmo3_indomain} have an average length of around 1K, this only indicates that HiLS-Attention has on-par ability to full attention in short contexts.
In contrast, the LongBench results in Tab.~\ref{tab:longbench_v1} can sufficiently demonstrate the long-context capability of HiLS-Attention. Even with the extrapolation enhancement provided by YaRN~\cite{peng2024yarn}, there remains a noticeable gap on LongBench compared with the training-free HiLS-Attention.

\section{Analysis}

\subsection{Inference Efficiency of HiLS-Attention}
We benchmark the end-to-end inference latency of HiLS-Attention against full attention in the SGLang~\cite{zheng2024sglangefficientexecutionstructured} inference engine on a single NVIDIA H800 GPU. 
To ensure an apple-to-apple comparison, both backends share the same Triton attention-kernel infrastructure and differ only in the attention algorithm. In particular,  the full-attention baseline is run with the same Triton kernels rather than a vendor-optimized paged-attention implementation. We use the 345M-scale architecture of Sec.~\ref{sec:small_scale_exp} with chunk size 64, top-$k=32$, and a 512-token sliding window. We set the paging granularity of the KV cache equal to the chunk size, and decode in a single stream (batch size 1) in bf16. We report warm-run prefill latency for the full input and the median per-token decode latency with CUDA graphs enabled.

\begin{figure}
    \centering
    \includegraphics[width=1\linewidth]{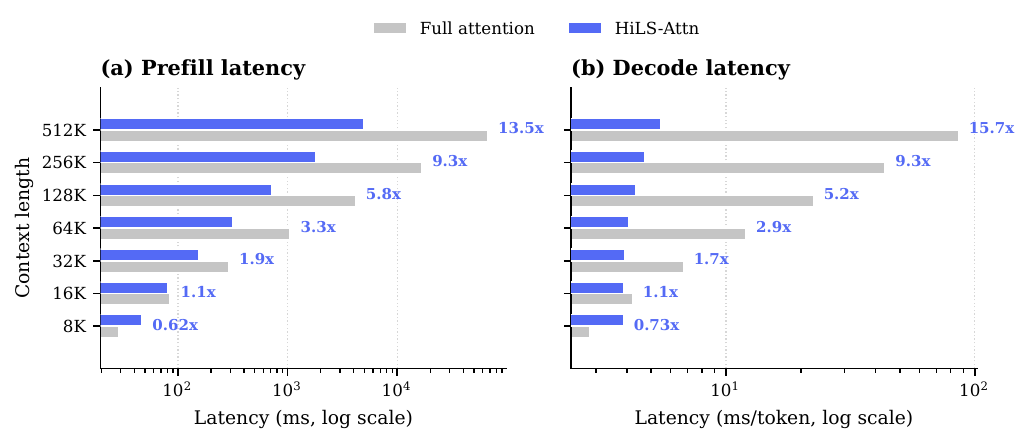}
     \caption{Inference latency of HiLS-Attention vs.\ full attention at the 345M scale on a single NVIDIA H800 (batch size 1, bf16, chunk size 64, top-$k=32$, 512-token sliding window). Prefill is the warm-run latency for the full input; decode is the median per-token latency with CUDA graphs. Speedup is full attention $/$ HiLS-Attention ($>1$ means HiLS-Attention is faster). HiLS-Attention reaches parity at $\sim$16K and is $13.5\times$/$15.7\times$ faster (prefill/decode) at 512K.}
    \label{fig:efficiency}
\end{figure}

As shown in Fig.~\ref{fig:efficiency}, the cost of HiLS-Attention is governed by its fixed retrieval budget ($K\times\text{chunk}=2048$ tokens) plus the local sliding window, rather than by the full context length. Consequently its prefill latency grows \emph{near-linearly} with context length while full attention grows quadratically. Its per-token decoding latency stays \emph{effectively constant} ($\mathcal{O}(1)$) while full attention grows linearly with the KV-cache length. The latency curves of the two methods cross over at roughly 16K tokens: below this length, the cost of top-$k$ retrieval is not yet amortized, so full attention is faster; at and beyond 16K, HiLS-Attention reaches parity on prefill and is already faster on decoding. Past the crossover the gap widens rapidly with length\textemdash at 512K context HiLS-Attention is $13.5\times$ faster in prefill (5.0\,s vs.\ 67.0\,s) and $15.7\times$ faster per decode step (5.5\,ms vs.\ 85.9\,ms)\textemdash  making HiLS-Attention substantially more practical than full attention for long-context serving while matching its quality (Sec.~\ref{sec:small_scale_exp}).

\subsection{Chunk Overlap}
We next validate a key empirical assumption behind the HiLS-Attention kernel: adjacent
autoregressive queries retrieve highly overlapping chunks. This is especially important for MHA
models, where GQA-style head grouping, as used in NSA, is unavailable to enlarge the GEMM dimension
for Tensor Cores. HiLS-Attention therefore adopts a \emph{one-load-multiple-compute} strategy: it
groups $M$ adjacent queries, loads the union of their retrieved chunks once, and computes attention
for all $M$ queries over this shared set. The union may introduce a few invalid query--chunk pairs,
but it improves data reuse, reduces memory traffic, and better utilizes Tensor Cores.

For a block of $M>1$ adjacent queries, let $\mathcal{I}_m$ denote the top-$K$ retrieved chunks of the
$m$-th query. We measure the normalized chunk-overlap ratio as
\begin{equation}
    \mathrm{Overlap}
    =
    \frac{M K - \left|\bigcup_{m=1}^{M}\mathcal{I}_m\right|}
    {(M-1)K},
\end{equation}
where $\mathrm{Overlap}=1$ means perfect overlap and $\mathrm{Overlap}=0$ means no overlap. We
evaluate the Olmo3-7B continued-pretraining checkpoint on pretraining sequences with chunk size 64
and top-$K=32$. We use $M=16$ for tail-block loading from 4K to 64K contexts, and sweep
$M\in\{2,4,8,16,32,64\}$ for the overlap ratio, with $M=1$ included as the single-query baseline.

\begin{figure}[t]
    \centering
    \includegraphics[width=0.95\linewidth]{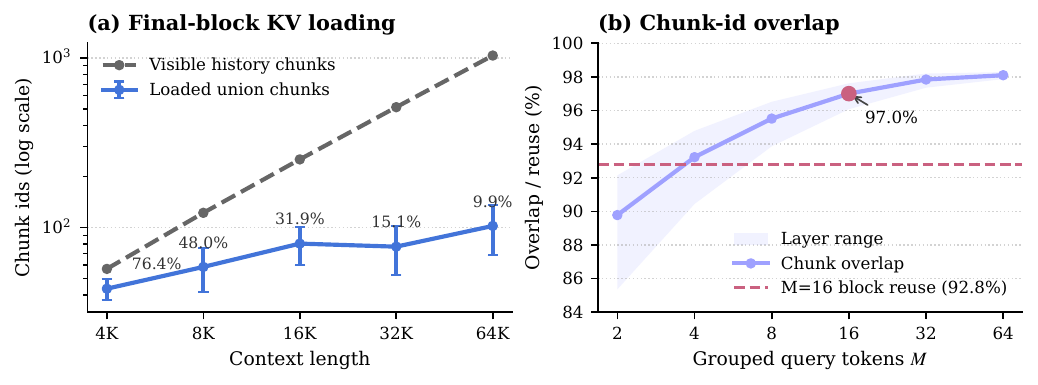}
    \caption{Chunk-id overlap among adjacent query tokens. Left: loaded union size for the final $M=16$ queries versus the visible historical chunks; percentages denote loaded fractions. Right: normalized overlap as group size $M$ increases, with the dashed line showing inter-block reuse at $M=16$. Error bars show standard deviation across HiLS layers and heads, and shaded regions show the layer-wise min--max range.}
    \label{fig:chunk_overlap}
\end{figure}

Fig.~\ref{fig:chunk_overlap} shows strong chunk sharing among adjacent queries. For the final
$M=16$ queries, the visible historical pool grows from 57 to 1032 chunks as context increases from
4K to 64K, yet the loaded union grows only from 43.6 to 102.1 chunks, reducing the loaded fraction
from 76.4\% to 9.9\%. Moreover, 92.8\% of chunks in a query block already appeared in the previous
block on average, further supporting adjacent-query packing and one-load-multiple-compute.

\section{Related Work}
Existing sparse attention methods can be broadly categorized into token-wise~\cite{liu2025deepseek} and block-wise approaches.
We focus on block-wise sparse attention, as it can potentially support native sparse training, which is essential for infinitely long context modeling. 
We further distinguish block-wise methods by whether their chunk summaries are parameterized. This distinction is important because chunk selection relies on summaries as proxies for chunk relevance, and their fidelity directly affects whether relevant chunks can be accurately retrieved.
\subparagraph{None-parametric summaries.}
Many block-wise sparse attention methods, including NSA~\cite{yuan-etal-2025-native}, MoBA~\cite{DBLP:journals/corr/abs-2502-13189}, use mean-pooled chunk representations for efficient chunk selection. DashAttention~\cite{huang2026dashattentiondifferentiableadaptivesparse} introduces a learnable gate based on InfLLM v2~\cite{zhao2025infllmv2densesparseswitchableattention}. Nevertheless, as shown in Tab.~\ref{tb:345M_main_ruler}, these methods fail to achieve perfect needle retrieval even in the training length, suggesting inaccurate chunk selection.
SeerAttention~\cite{gao2025seerattentionlearningintrinsicsparse} also uses mean pooling as the chunk summary, but the chunk selection is distilled from full-attention. Since the upper bound of a distillation-based method is full attention, we do not include it in our experiments.

\subparagraph{Parametric summaries.}
Landmark Attention (LMK-Attn)~\cite{mohtashami2023random} first introduced landmark tokens to learn chunk summaries, but requires dense attention during training and applies sparsification only at inference. 
Moreover, its significant higher perplexity than the full-attention baseline makes it difficult to match full attention on downstream tasks.
HSA series~\cite{hu2026hardware,hu2026everytoken} enables sparse training via specialized kernel design and empirically shows strong length generalization~\cite{leng2026understanding} by combining NoPE-based HSA with RoPE-based SWA. 
However, HSA integrates with SWA as a cross-attention module, which introduces substantial parameter overhead. Furthermore, after prolonged training, the perplexity (PPL) becomes dominated by SWA, as demonstrated in Tab.~\ref{tb:345M_main_ppl}.

Compared with LMK-Attn, our work enables native sparse training and provides a more accurate estimate of chunk mass. LMK-Attn lags behind the full-attention baseline because it uses only the landmark token key as the chunk summary. In contrast, our method, grounded in Proposition~\ref{prop:logsumexp}, achieves substantially better perplexity and extrapolation. Inspired by HSA’s use of NoPE, we adopt HoPE~\cite{chen-etal-2025-hope} to preserve in-domain positional awareness while enabling ultra-long context extrapolation. We provide an intuitive comparison with other sparse attention methods across multiple dimensions in Tab.~\ref{tab:attention_comparison}.

\newcommand{\diff}[1]{\textcolor{softred}{#1}}
\newcommand{\hilight}[1]{\textcolor{softgreen}{\textbf{#1}}}

\begin{table}[htb!]
\centering
\small
\setlength{\tabcolsep}{6pt}
\renewcommand{\arraystretch}{1.18}

\begin{threeparttable}
\begin{tabular}{@{}lccccc}
\toprule
\textbf{Method}
& \makecell{\textbf{Perfect}\\\textbf{NIAH}}
& \makecell{\textbf{Full}\\\textbf{QK}}
& \textbf{Extrapolation}
& \makecell{\textbf{Training}\\\textbf{Strategy}}
& \makecell{\textbf{Supports}\\\textbf{CPT}} \\
\midrule

Full-Attention
& Yes
& \diff{Yes}
& \diff{$<2\times$}
& End-to-End
& Yes \\

Seer-Attention
& Yes
& \diff{Yes}
& \textemdash
& \diff{Distillation}
& Yes \\

NSA
& \diff{No}
& No
& \diff{$<2\times$}
& End-to-End
& Yes \\

DSA
& Yes
& \diff{Yes}
& \textemdash
& \diff{Distillation}
& Yes \\

Dash-Attention
& \diff{No}
& No
& \diff{$<2\times$}
& End-to-End
& Yes \\

LMK-Attn
& Yes
& \diff{Yes}
& \diff{$< 32\times$}
& End-to-End
& Yes \\

HSA
& \diff{No}
& No
& $>64\times$
& End-to-End
& \diff{No} \\

\rowcolor{blue!10}
{HiLS-Attention}
& {Yes}
& {No}
& {$>64\times$}
& {End-to-End}
& {Yes} \\

\bottomrule
\end{tabular}

\end{threeparttable}
\vspace{0.1cm}
\caption{Comparison of sparse-attention variants in terms of in-context retrieval capability, require full QK computation during training, extrapolation ability, training strategy, and compatibility with CPT. The weaknesses of each method are highlighted in \diff{red}.}
\label{tab:attention_comparison}
\end{table} 
\section{Discussion \& Conclusion}
This work introduces HiLS-Attention, a sparse attention mechanism featuring end-to-end retrieval learning and native sparse training.
Our experiments show that HiLS-Attention exhibits strong length extrapolation ability. This raises a natural question: what is the source of such extrapolation ability, and is it only reflected under out-of-distribution context lengths?
\paragraph{Our hypothesis}
We argue that stronger extrapolation essentially stems from more accurate in-context retrieval, which is enabled by the compression-and-retrieval inductive bias of HiLS-Attention.
During compression, token-level noise can partially be canceled out, while shared semantic signals are preserved. This  leads  to higher-quality retrieval key representations.
The advantage of HiLS-Attention for in-domain retrieval becomes especially evident under longer contexts.
As long-context capability becomes increasingly important for modern LLMs and agent models, our findings provide useful insights for improving long-context modeling.

\paragraph{Toward infinite context modeling}
Although HiLS-attention achieves ultra-long context capability mainly through extrapolation in this work, it also provides a practical route toward training with ultra-long contexts. Its key advantage lies in its native sparsity during training and length generalization. Unlike full-attention or distillation-based methods~\cite{liu2025deepseek}, which score all token-level pairs with quadratic complexity $O(L^2)$, HiLS first compresses each chunk into a lightweight summary via intra-chunk attention. Each query token then retrieves from only $L/S$ chunk summaries, resulting in a per-token retrieval cost of $O(L/S)$ and a sequence-level retrieval cost of $O(L^2/S)$. 

Once the relevant chunks are retrieved, the actual token-level attention is restricted to the selected chunks and a local window, yielding a per-token attention cost of $O(KS)$ and a total attention cost of $O(LKS)$. Thus, the only remaining superlinear component is retrieval, which can be further reduced to approximately $O(L\log(L/S))$ with approximate nearest-neighbor search methods like FAISS~\cite{11202651}.

\paragraph{Limitation and Future works}
HiLS-Attention does not support context parallelism yet, so its effectiveness under larger training context lengths remains to be fully validated. In addition, unselected chunks receive no gradient updates. The underlying mechanism by which Q-Cal improves extrapolation and in-domain performance is still not fully understood. We leave further validation and improvement to future work.

\clearpage
{
	\small
	\bibliographystyle{unsrtnat}
	\bibliography{references}
}

\clearpage
\appendix
\section{Justification of Equation~\ref{eq:two_regime}}\label{apdx:logsumexp_two_regime}
Let $S=|\mathcal{T}_c|$ and write $s_j=s_{i,j}$.
When the logits are nearly uniform, let
$
s_j=\bar{s}+\delta_j
$
where
$
\bar{s}=\frac{1}{S}\sum_j s_j
$
and
$
\sum_j\delta_j=0
$.
Then
\begin{align}
\log\sum_j \exp(s_j)
&= \bar{s}+\log\sum_j\exp(\delta_j) \\
&= \bar{s}+\log\left(S+\frac{1}{2}\sum_j\delta_j^2+O(\|\delta\|^3)\right) \\
&= \bar{s}+\log S+O\left(\frac{1}{S}\sum_j\delta_j^2\right).
\end{align}
Hence, for small within-chunk logit variance,
$
\log\sum_j\exp(s_j)\approx \operatorname{mean}_j(s_j)+\log S
$.

When one logit dominates, let
$
M=\max_j s_j
$
and
$
j^\star=\arg\max_j s_j
$.
Then
\begin{align}
\log\sum_j\exp(s_j)
&= M+\log\left(1+\sum_{j\neq j^\star}\exp(s_j-M)\right).
\end{align}
If $M-s_j$ is large for all $j\neq j^\star$, the second term is negligible, and therefore
$
\log\sum_j\exp(s_j)\approx M=\max_j s_j
$.
\section{Proof of Proposition~\ref{prop:logsumexp}}\label{appdx:proof_of_logsumexp}

Given $\boldsymbol{q} \in \mathbb{R}^{d}$ and $\mathbf{K} \in \mathbb{R}^{S\times d}$, where $S$ is the block size, we explain how to estimate the Log-Sum-Exp function \footnote{We omit the scaling factor $s=\frac{1}{\sqrt{d}}$ here for simplicity} $f(\boldsymbol{q}) = \log\sum_{j=1}^{S} \exp({\boldsymbol{q}\mathbf{K}_j})$ through the first-order Taylor expansion at $\boldsymbol{q}_c$:
\begin{equation}
f(\boldsymbol{q}) \approx f(\boldsymbol{q}_c) + \nabla f(\boldsymbol{q}_c)^{\top} (\boldsymbol{q} - \boldsymbol{q}_c),\label{eq:f_q_taylor}
\end{equation}
where we have:
\begin{align}
\nabla f(\boldsymbol{q}_c) &= \nabla_{\boldsymbol{q}_c} \Big(\log\sum_{j=1}^{S} \exp({\boldsymbol{q}_c^\top\mathbf{K}_j} )\Big) \\
&= \frac{1}{\sum_{j=1}^{S}\exp({\boldsymbol{q}_c^\top\mathbf{K}_j})}\nabla_{\boldsymbol{q}_c}\Big(\sum_{j=1}^{S} \exp({\boldsymbol{q}_c^\top\mathbf{K}_j} )\Big) \nonumber \\
&= \frac{1}{\sum_{j=1}^{S}\exp({\boldsymbol{q}_c^\top\mathbf{K}_j})}\Big(\sum_{j=1}^{S} \exp({\boldsymbol{q}_c^\top\mathbf{K}_j} )\mathbf{K}_j\Big) \nonumber\\
&= \sum_{j=1}^{S} \underbrace{\frac{\exp({\boldsymbol{q}_c^\top\mathbf{K}_j})}{\sum_{j=1}^{S}\exp({\boldsymbol{q}_c^\top\mathbf{K}_j})}}_{p_j}
\mathbf{K}_j \label{eq:nabla_fq_pj} \\
&= \boldsymbol{k}'_c.
\end{align}
Thus, we can re-formalize Eq.\ref{eq:f_q_taylor} as follows:
\begin{align}
    f(\boldsymbol{q}) &\approx \log\sum_{j=1}^{S} \exp({\boldsymbol{q}_c^\top\mathbf{K}_j} ) + {\boldsymbol{k}'_c}^{\top}(\boldsymbol{q} - \boldsymbol{q}_c) \\
    &\overset{(i)}{=} {\boldsymbol{k}'_c}^{\top}\boldsymbol{q} + \Big(\log\sum_{j=1}^{S} \exp({\boldsymbol{q}_c^\top\mathbf{K}_j} ) - {\boldsymbol{k}'_c}^\top \boldsymbol{q}_c \Big) \nonumber \\
    & \overset{(ii)}{=} {\boldsymbol{k}'_c}^{\top}\boldsymbol{q} + \Big(\log\sum_{j=1}^{S} \exp({\boldsymbol{q}_c^\top\mathbf{K}_j} ) - \sum_{j=1}^{S} p_j \boldsymbol{q}_c^\top\mathbf{K}_j\Big) \nonumber\\
    & \overset{(iii)}{=} {\boldsymbol{k}'_c}^{\top}\boldsymbol{q} + \Big(\underbrace{\log\sum_{j=1}^{S} \exp({\boldsymbol{q}_c^\top\mathbf{K}_j} )}_{\textrm{Constant}} - \sum_{j=1}^{S} p_j \big(\log p_j + \underbrace{\log\sum_{j=1}^S\exp({\boldsymbol{q}_c^\top\mathbf{K}_j})}_{\textrm{Constant}}\big)\Big) \nonumber\\
    & \overset{(iv)}{=} {\boldsymbol{k}'_c}^{\top}\boldsymbol{q}  - \sum_{j=1}^{S} p_j\log p_j 
\end{align}
where $(i)$ re-orders the equation with the second term be a bias value, $(ii)$ re-uses Eq. \ref{eq:nabla_fq_pj}, $(iii)$ re-formalizes $\boldsymbol{q}_c^\top\mathbf{K}_j$ as follows:
\begin{align}
    \boldsymbol{q}_c^\top\mathbf{K}_j &= \log\exp(\boldsymbol{q}_c^\top\mathbf{K}_j) \nonumber \\
    &= \log\underbrace{\frac{\exp({\boldsymbol{q}_c^\top\mathbf{K}_j})}{\sum_{j=1}^{S}\exp({\boldsymbol{q}_c^\top\mathbf{K}_j})}}_{p_j} + \log\sum_{j=1}^{S}\exp({\boldsymbol{q}_c^\top\mathbf{K}_j}), \label{eq:qk_new_form}
\end{align}
and $(iv)$ cancels the constant terms. The final objective is the same as Eq. \ref{eq:attn_lhsa}

\section{HiLS-Atention in GQA}\label{appdx:hils-attn-gqa}
Considering an arbitrary group in GQA, given $\mathbf{q}_i \in \mathbb{R}^{G \times d}$ and $\mathbf{k}_i, \mathbf{v}_i \in \mathbb{R}^{d}$, where $G$ query heads share one KV head.
Note that in GQA mode, the shape of $\mathbf{k}'_c$ is consistent with that of the query, i.e., $\mathbf{k}'_c \in \mathbb{R}^{G \times d}$.
We have:
\begin{equation}
\label{eq:hils_gqa}
\small
\begin{aligned}
    &s_{i,j}^h = \frac{(\mathbf{q}_i^h)^\top \mathbf{k}_j}{\sqrt{d}},\quad
    \hat{s}^h_{i,c} = \frac{(\hat{\mathbf{q}}_i^h)^\top \mathbf{k}'_c}{\sqrt{d}} + b'_{c}, \quad \hat{s}_{i,c} = \underset{h}{\text{argmax}}(\hat{s}^h_{i,c}), \\
    & Z^h_{i,\text{swa}}=\sum_{j\in \mathcal{T}_c}{\exp(s^h_{i,j})},\quad \hat{Z}^h_{i,c}=\exp(\hat{s}_{i,c}^h),\\
    &\bm{\mathcal{I}}_i = \left\{ c \in \mathcal{C}_i \mid \operatorname{rank}_{\downarrow}(\hat{s}_{i,c}) < K \right\}, \quad \hat{\mathbb{Z}}^h_i = \sum_{c \in \bm{\mathcal{I}}_i} \hat{Z}^h_{i,c} + Z^h_{i,\text{swa}}, \\
    &w^h_{i,j} = \begin{cases}
    \displaystyle \frac{\exp(s^h_{i,j})}{Z^h_{i,c(j)}} \cdot \frac{\hat{Z}^h_{i,c(j)}}{\hat{\mathbb{Z}}^h_i} & ,\text{if } j < \ell(i) \text{ and } c(j) \in \bm{\mathcal{I}}_i \\[1.2em]
    \displaystyle \frac{\exp(s^h_{i,j})}{\hat{\mathbb{Z}}^h_i} & ,\text{if } \ell(i) \le j \le i \\[1.2em]
    0 & ,\text{otherwise}
    \end{cases}, \quad \mathbf{o}^h_i = \sum_j w^h_{i,j}\mathbf{v}^h_j.
\end{aligned}
\end{equation}

\section{Hyper-parameters}\label{appdx:hyper-params}
\begin{table}[h!]
\centering
\resizebox{0.65\textwidth}{!}{
\begin{tabular}{l|c|c|c}
\toprule
Hyperparameter & Small scale & Medium scale & Large scale\\
\midrule
Parameters & 345M & 1.4B & 7B\\
Layers & 16 & 22 & 32\\
Hidden size & 1024 & 2048 & 4096\\
FFN size & 4096 & 5632 & 11008\\
Q/KV heads & 16/2 & 32/4 & 32/32\\
HiLS layers & every layer & every layer & every 4 layers\\
SWA window & 512 & 512 & 512\\
Chunk size & 64 & 64 & 64\\
HiLS top-$K$ & 32 & 32 & 32\\
LoRA-Q bottleneck & 64 & 128 & 256\\
Vocab size & $100278 + 1$ & $100278 + 1$ & $100278 + 1$\\
\bottomrule
\end{tabular}
}
\vspace{0.1cm}
\caption{Model hyperparameters used in the small-, medium-, and large-scale experiments. The vocab size is the original OLMo3 vocabulary plus one additional landmark-token embedding. Chunk/Top-$k$ = 64/32 corresponds to retrieving 2048 tokens.}
\end{table}

\section{Training recipes}\label{apdx:training_recipe}
\paragraph{Small-scale models.}
The training data is based on OLMo/Dolma pre-training corpora~\citep{allenai_dolma3_mix_6t}, mixed with 5\% RULER-style synthetic examples to provide the instruction-following ability needed to answer RULER-style retrieval queries. The training context length is 8K, the global batch size is 128, and the models are trained for 30K optimizer steps, corresponding to approximately 30B training tokens. We use the AdamW optimizer with a constant learning-rate schedule of \(3\times10^{-4}\). Unless otherwise specified, all models share the same tokenizer, training data, training length, and optimization configuration. We evaluate the small-scale models from two perspectives: perplexity is used to measure language modeling ability, while RULER~\citep{hsieh2024ruler} is used to evaluate long-context random access and order-aware retrieval capability.

\paragraph{7B continue pre-training recipe.} For continue pretraining, we use the OLMo3 pretraining corpus distribution. Specifically, we train from the tokenized OLMo3 500B-token~\citep{allenai_dolma3_mix_6t} pretraining pool and sample approximately 50B tokens for adaptation by drawing 8K sequences from this pool. We set the maximum sequence length to 8192, the global batch size to 512, and train for 13K optimizer steps. The learning rate warms up for 1K steps to $2\times10^{-4}$ and then follows a cosine decay to $2\times10^{-5}$.
The original OLMo3-Base checkpoint is not instruction-tuned and therefore lacks the instruction-following ability needed to answer RULER-style retrieval queries at test time.
To make long-context retrieval comparisons meaningful after CPT, we mix on-the-fly synthesized RULER examples into the training stream at a 5\% rate: with this probability, each sampled 8K pretraining sequence is converted into one of the three evaluation tasks---single needle-in-a-haystack, multi-query, and variable tracking---while the remaining 95\% of samples stay as plain corpus text.
All continued-pretraining models in Table~\ref{tab:olmo3_indomain}, including Olmo3-512swa-CPT and the HiLS-Attn variants, follow this recipe; the \textbf{Olmo3-Base} row reports the original checkpoint without RULER-augmented CPT.

\section{Downstream Evaluation Details}\label{apdx:benchmarks}
\begin{itemize}[leftmargin=*,itemsep=0.2em,topsep=0.3em]
    \item LAMBDA~\cite{paperno2016lambada}
    \item HellaSwag~\cite{zellers2019hellaswag}
    \item PIQA~\cite{bisk2020piqa}
    \item WinoGrande~\cite{sakaguchi2021winogrande}
    \item OpenBookQA~\cite{mihaylov2018can}
    \item ARC-challenge~\cite{clark2018think}
    \item ARC-esay: an easy subset of ARC-challenge
    \item MMLU~\cite{hendryckstest2021}
    \item GPQA~\cite{rein2024gpqa}
    \item BoolQ~\cite{clark2019boolq}
    \item RACE~\cite{lai-etal-2017-race}
    \item CMATH~\cite{wei2023cmathlanguagemodelpass}
    \item GSM8K~\cite{cobbe2021gsm8k}
    \item CRUX~\cite{gu2024cruxevalbenchmarkcodereasoning}
    \item HumanEval+~\cite{evalplus,evalperf}
    \item MBPP+~\cite{evalplus,austin2021programsynthesislargelanguage}
\end{itemize}

\section{Perplexity at Different CPT Steps}\label{apdx:ppl_in_cpt}
Table~\ref{tab:ppl_in_cpt} reports in-domain perplexity on the OLMo3 pretraining corpus at 128, 512, and 8K context lengths for CPT checkpoints from 6K to 13K.
We define the signed gap as $\Delta\mathrm{PPL}=\mathrm{PPL}_{\mathrm{HiLS\text{-}Attn}}-\mathrm{PPL}_{\mathrm{Olmo3\text{-}512swa\text{-}CPT}}$ (positive = HiLS worse).
The gap is largest at short context early in training (e.g., $+0.23$ at 128 tokens, 6K) and shrinks with more steps (down to $+0.02$--$+0.06$ at 128 and $+0.02$--$+0.03$ at 512 by 10K--13K), while staying within $\pm 0.01$ at 8K throughout.

\begin{table}[h!]
    \centering
    \small
    \setlength{\tabcolsep}{14pt}
    \renewcommand{\arraystretch}{1.15}
    \begin{tabular}{lccc}
    \toprule
    \multirow{2}{*}{CPT Step}
    & \multicolumn{3}{c}{$\Delta\mathrm{PPL}$}
    \\
    \cmidrule(lr){2-4}
    & \makecell{128\\tokens}
    & \makecell{512\\tokens}
    & \makecell{8K\\tokens}
    \\
    \midrule
    6K  & $+0.23$ & $+0.07$ & $+0.01$ \\
    7K  & $+0.13$ & $+0.07$ & $+0.00$ \\
    8K  & $+0.19$ & $+0.06$ & $+0.01$ \\
    9K  & $+0.02$ & $+0.04$ & $+0.00$ \\
    10K & $+0.06$ & $+0.02$ & $-0.01$ \\
    11K & $+0.06$ & $+0.02$ & $+0.00$ \\
    12K & $+0.05$ & $+0.02$ & $+0.00$ \\
    13K & $+0.05$ & $+0.03$ & $+0.01$ \\
    \bottomrule
    \end{tabular}
    \caption{Signed perplexity gap across CPT steps:
    $\Delta\mathrm{PPL}=\mathrm{PPL}_{\mathrm{HiLS\text{-}Attn\text{-}HoPE\text{-}LoRA}}-\mathrm{PPL}_{\mathrm{Olmo3\text{-}512swa\text{-}CPT}}$.
    Positive $\Delta$ means HiLS-Attn has higher (worse) perplexity.}
    \label{tab:ppl_in_cpt}
\end{table}

\section{Per-Step Results of the 1.4B Model}\label{apdx:1B_perstep}
We provide the full per-step perplexity (Table~\ref{tb:1B_ppl_steps}) and RULER
(Table~\ref{tb:1B_ruler_steps}) numbers of the 1.4B model across training steps.

\begin{table}[tb]
\caption{Perplexity for 1.4B model after 300B-token training with an 8K context length.}\label{tb:1B_ppl_steps}
\vspace{0.1cm}
\centering
\resizebox{0.65\textwidth}{!}{
\begin{tabular}{lccccccccc}
\toprule
 & \#steps & 64 & 128 & 512 & 8K & 32K & 128K & 512K \\
\midrule
\multirow{8}{*}{\rotatebox[origin=c]{90}{Full-Attn RoPE}}  &     20k &  33.78 & 25.94 & 17.56 & 4.63 & $>10^2$ \\
& 40k  & 31.01 & 23.81 & 16.08 & 4.39 & $>10^2$ \\
& 60k  & 29.47 & 22.61 & 15.30 & 4.29 & 5.87& $>10^2$ \\
& 80k  & 28.21 & 21.72 & 14.69 & 4.20 & 5.49 & $>10^2$\\
& 100k & 27.25 & 20.92 & 14.16 & 4.12 & 6.09 & $>10^2$\\
& 120k & 26.60 & 20.40 & 13.79 & 4.05 & 6.56 & $>10^2$\\
& 140k & 26.18 & 20.05 & 13.56 & 4.02 & 8.28 & $>10^2$\\
& 143k & 26.17 & 20.05 & 13.56 & 4.02 & 8.60 & $>10^2$\\

\midrule
\multirow{8}{*}{\rotatebox[origin=c]{90}{HiLS-Attn HoPE}} &     20k & 33.88 & 26.03 & 17.58 & 4.63 & 4.10 & 4.69 & 7.08 \\
& 40k  &  31.10 & 23.84 & 16.11 & 4.40  & 4.03 & 4.68 & 6.62 \\
& 60k  & 29.47& 22.55&  15.30& 4.29 & 4.17 & 5.65 & 8.49 \\
& 80k  &  28.20 &  21.67 & 14.70 & 4.20 & 3.98 & 5.30 &  8.35\\
& 100k & 27.40 & 20.98 & 14.19 & 4.12 & 4.00 & 5.38 & 8.24 \\
& 120k & 26.63 & 20.39 & 13.79 & 4.05 & 3.99 & 5.43 & 8.48 \\
& 140k & 26.26 & 20.08 & 13.57 & 4.02 & 4.08 & 5.54 & 8.53 \\
& 143k & 26.21 & 20.03 & 13.55 & 4.02 & 4.08 & 5.54 & 8.58 \\

\bottomrule
\end{tabular}
}
\end{table}

\begin{table}[tb]
\centering
\caption{RULER results of the 1.4B model at different training steps during 300B-token training.}\label{tb:1B_ruler_steps}
\setlength{\tabcolsep}{3pt} 
\vspace{0.1cm}
\resizebox{0.9\textwidth}{!}{%
\begin{tabular}{l|c*{5}{|ccc}}
\toprule
\multirow{2}{*}{} & \multirow{2}{*}{Steps}
& \multicolumn{3}{c|}{8K}
& \multicolumn{3}{c|}{16K}
& \multicolumn{3}{c|}{32K}
& \multicolumn{3}{c|}{128K}
& \multicolumn{3}{c}{512K} \\

& & S-N & MK-MQ & VT
& S-N & MK-MQ & VT
& S-N & MK-MQ & VT
& S-N & MK-MQ & VT
& S-N & MK-MQ & VT \\
\midrule

\multirow{8}{*}{\rotatebox[origin=c]{90}{Full-Attn RoPE}}      
 & 20k & 99 & 58 & 71 & 4 & 0 & 0 & 0 & 0 & 0 & 0 & 0 & 0 & 0 & 0 & 0 \\
 & 40k & 100 & 99 & 82 & 35 & 0 & 0 & 0 & 0 & 0 & 0 & 0 & 0 & 0 & 0 & 0 \\
 & 60k & 100 & 99 & 86 & 33 & 15 & 11 & 4 & 0 & 0 & 0 & 0 & 0 & 0 & 0 & 0\\
 & 80k & 100 & 99 & 94 & 44 & 19 & 11 & 5 & 0 & 0 & 0 & 0 & 0 & 0 & 0 & 0\\
 & 100k & 100 & 99 & 91 & 35 & 14 & 4 & 2 & 0 & 0 & 0 & 0 & 0 & 0 & 0 & 0\\
 & 120k & 100 & 98 & 95 & 27 & 2 & 3 & 3 & 0 & 0 & 0 & 0 & 0 & 0 & 0 & 0\\
 & 140k & 100 & 97 & 95 & 27 & 0 & 3 & 2 & 0 & 0 & 0 & 0 & 0 & 0 & 0 & 0\\
 & 143k & 100 & 96 & 94 & 21 & 0 & 2 & 3 & 0 & 0 & 0 & 0 & 0 & 0 & 0 & 0\\
\midrule
\multirow{8}{*}{\rotatebox[origin=c]{90}{HiLS-Attn HoPE}}             
 & 20k & 100 & 100 & 77 & 100 & 98 & 66 & 99 & 95 & 61 & 99 & 93 & 56 & 79 & 90 & 60 \\
 & 40k & 100 & 98 & 75 & 100 & 99 & 60 & 100 & 95 & 64 & 99 & 95 & 65 & 85 & 91 & 64 \\
 & 60k & 100 & 99 & 84 & 100 & 99 & 67 & 100 & 97 & 71 & 94 & 98 & 59 & 67 & 94 & 54 \\
 & 80k & 100 & 100 & 85 & 100 & 98 & 74 & 100 & 96 & 77 & 99 & 91 & 68 & 76 & 84 & 50 \\
 & 100k & 100 & 100 & 90 & 100 & 100 & 71 & 100 & 96 & 76 & 94 & 98 & 71 & 64 & 92 & 52 \\
 & 120k & 100 & 99 & 90 & 100 & 99 & 70 & 100 & 99 & 76 & 98 & 97 & 70 & 87 & 97 & 64 \\
 & 140k & 100 & 98 & 88 & 100 & 96 & 76 & 100 & 97 & 81 & 96 & 95 & 69 & 75 & 96 & 60 \\
 & 143k & 100 & 98 & 91 & 100 & 99 & 76 & 100 & 99 & 73 & 99 & 99 & 68 & 92 & 99 & 60 \\

\bottomrule
\end{tabular}%
}
\end{table}

\end{document}